%% file: Main.tex
\documentclass[10pt,twocolumn,letterpaper]{article}

\usepackage[pagenumbers]{cvpr} 

\usepackage{graphicx}
\usepackage{amsmath}
\usepackage{amssymb}
\usepackage{booktabs}
\usepackage{times}
\usepackage{epsfig}
\usepackage{multirow}
\usepackage{paralist}
\newcommand{\ours}{ConST-CL\xspace} 

\usepackage[colorlinks,urlcolor=blue,bookmarks=true,linkcolor=blue,citecolor=blue,urlcolor=blue,pagebackref=true]{hyperref}

\usepackage[british,american]{babel}

\usepackage[dvipsnames]{xcolor}

\definecolor{citecolor}{HTML}{0071BC}
\definecolor{linkcolor}{HTML}{ED1C24}

\usepackage[capitalize]{cleveref}
\crefname{section}{Sec.}{Secs.}
\Crefname{section}{Section}{Sections}
\Crefname{table}{Table}{Tables}
\crefname{table}{Tab.}{Tabs.}

\def\cvprPaperID{5430} 
\def\confName{CVPR}
\def\confYear{2022}

\DeclareMathOperator*{\argmin}{argmin}

\begin{document}

\title{Contextualized Spatio-Temporal Contrastive Learning with Self-Supervision}

\author{
Liangzhe Yuan$^{1}$\qquad
Rui Qian$^{1,2}\thanks{Work done as a student researcher at Google}$\qquad
Yin Cui$^{1}$\qquad
Boqing Gong$^{1}$\qquad
\\
Florian Schroff$^{1}$\qquad
Ming-Hsuan Yang$^{1}$\qquad
Hartwig Adam$^{1}$\qquad
Ting Liu$^{1}$ \\[2mm]
$^{1}$Google Research \qquad $^{2}$Cornell University}
\maketitle

\begin{abstract}
Modern self-supervised learning algorithms typically enforce persistency of instance representations across views.
%
%
While being very effective on learning holistic image and video representations, such an objective becomes sub-optimal for learning spatio-temporally fine-grained features in videos, where scenes and instances evolve through space and time.
In this paper, we present Contextualized Spatio-Temporal Contrastive Learning (\ours) to effectively learn spatio-temporally fine-grained video representations via self-supervision.
%
We first design a region-based pretext task which requires the model to transform instance representations from one view to another, guided by context features.
Further, we introduce a simple network design that successfully reconciles the simultaneous learning process of both holistic and local representations.
%
%
We evaluate our learned representations on a variety of downstream tasks and show that \ours achieves competitive results on 6 datasets, including Kinetics, UCF, HMDB, AVA-Kinetics, AVA and OTB.
Our code and models will be available at \small{\url{https://github.com/tensorflow/models/tree/master/official/projects/const_cl}}.
\end{abstract}
\vspace{-10pt}

\begin{figure}
    \centering
    \subcaptionbox{\label{fig:task_a}}{\includegraphics[width=0.465\columnwidth]{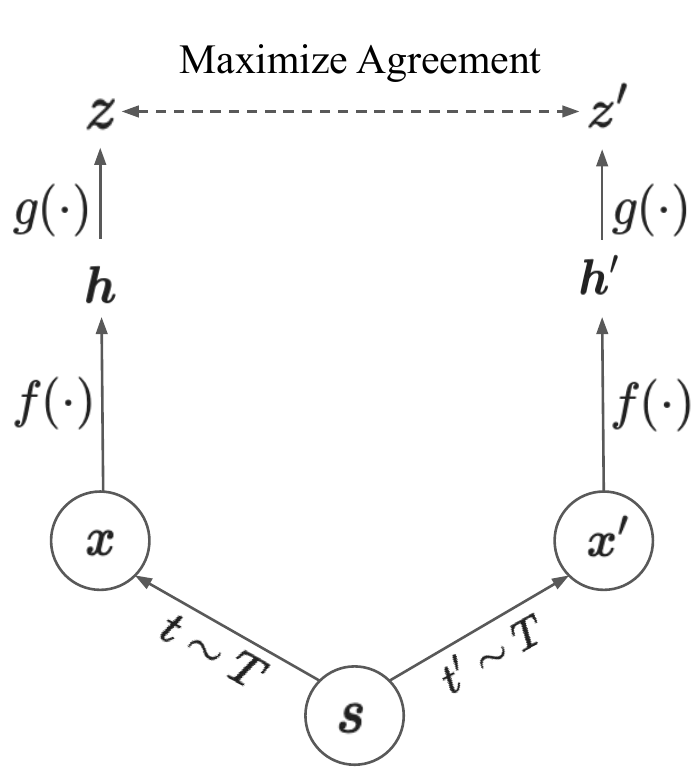}}%
    \hfill
    \subcaptionbox{\label{fig:task_b}}{\includegraphics[width=0.535\columnwidth]{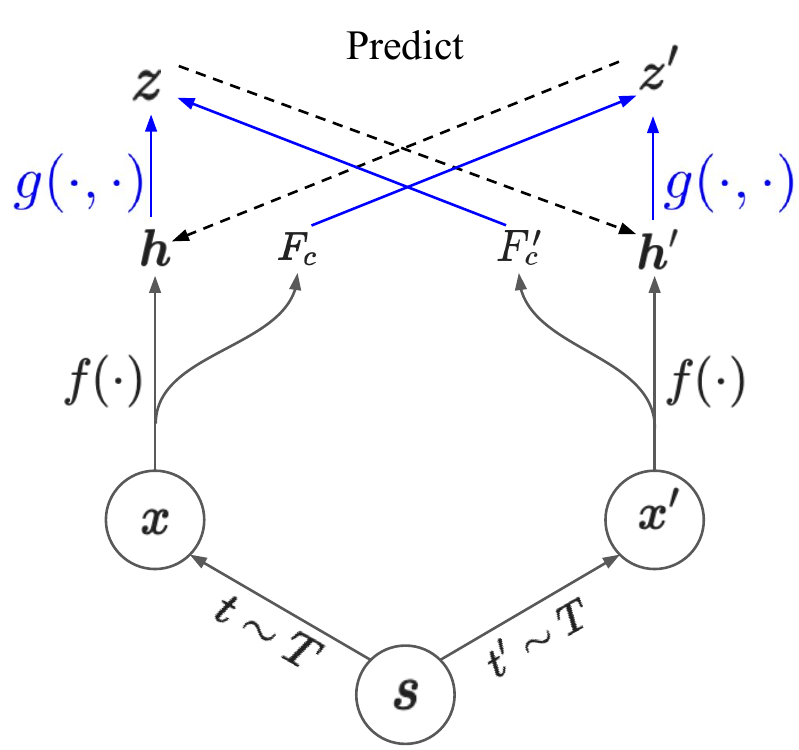}}%
    \vspace{-5pt}
    \caption{
    (a) A typical contrastive learning algorithm draws two augmented views $\{\boldsymbol{x},\boldsymbol{x^\prime}\}$ from one source $\boldsymbol{s}$ and trains an encoder network $f(\cdot)$ to construct representations $\boldsymbol{h}$ and $\boldsymbol{h'}$. 
    A projection function $g(\cdot)$ is trained to project representations into a shared space and to maximize the agreement between two views.
    (b) Contextualized Spatio-Temporal Contrastive Learning uses a binary projection function $g(\cdot,\cdot)$ to transform a representation $\boldsymbol{h}$ from one view to the other guided by context features $\boldsymbol{F_c^\prime}$ from the other view.
    %
    The contrastive objective encourages the transformed representation $\boldsymbol{z}$ to agree with its correspondence $\boldsymbol{h^\prime}$.
    }
    \label{fig:task}
    \vspace{-5pt}
\end{figure}
\input{1.Introduction.tex}

\begin{figure*}
    \centering
    \includegraphics[width=.95\textwidth]{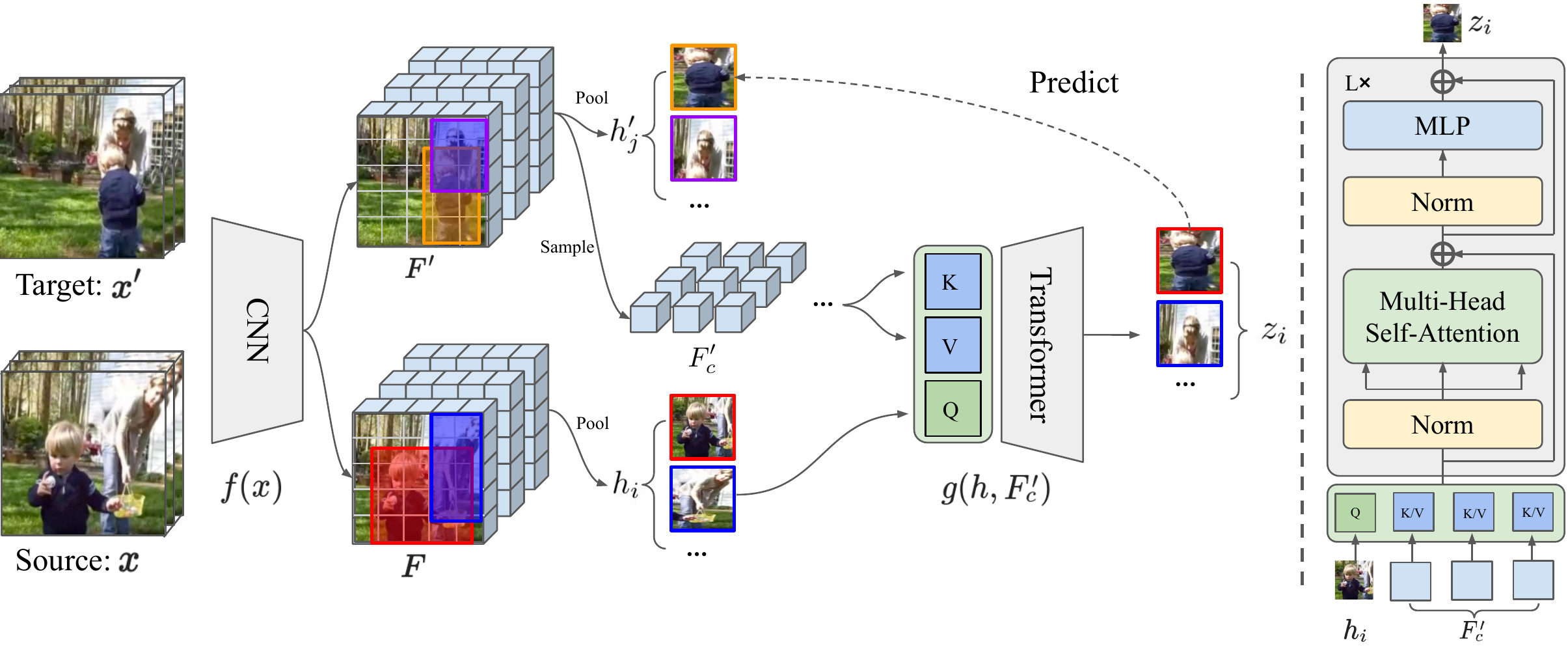}
    \vspace{-5pt}
    \caption{
    \textbf{Contextualized Spatio-Temporal Contrastive Learning.}
    Two spatio-temporally distant views are randomly sampled from one video and their dense representation feature maps $\{F,F^\prime\}$ are extracted by the base network $f(x)$. 
    Region features $\{h,h^\prime\}$ are pooled from respective dense feature maps by spatio-temporal ROIAlign and $F_c^\prime$ are a set of context features sampled from the dense feature maps $F^\prime$.
    The projection head $g(h, F_c^\prime)$ is learned to transform representations $h$ from one view to the other, guided by the context features $F_c^\prime$. We use the transformer~\cite{vaswani2017attention} architecture that takes region feature $h$ as Query and context features $F_c^\prime$ as Keys and Values.
    The InfoNCE loss is used to encourage the similarity between the reconstructed representations $z$ and their correspondences $h^\prime$.
    }
    \label{fig:method}
    \vspace{-5pt}
\end{figure*}

\input{2.RelatedWork.tex}
\input{3.Method.tex}
\input{4.Experiments.tex}
\input{5.Conclusion.tex}

\clearpage
{\small
\bibliographystyle{ieee_fullname}
\bibliography{egbib}
}

\clearpage
\begin{appendix}

\section{Model Architectures}
We illustrate the model architectures that are used in the \ours framework.
\subsection{Base network $\boldsymbol{f(\cdot)}$}
Table~\ref{tab:r3d_network} describes the base model architecture that is proposed for reconciling global and local training signals.
\input{Sup_Materials/r3d_table}

\subsection{\ours head $\boldsymbol{g(\cdot, \cdot)}$}
Table~\ref{tab:tx_network} describes the projection head we use for achieving the instance prediction task.
\input{Sup_Materials/tx_table}

\section{Region Generation}
In this section, we detail three options to generate region priors that we study for training \ours. 

\noindent \textbf{Random boxes.} 
For all of our related experiments, we randomly generate $8$ boxes on each frame. 
The boxes are constrained to have aspect ratio within $[0.5, 2]$ and size within $[0.1, 0.5]$ of the image size.
%

\noindent \textbf{Boxes from low-level image cues.}
%
We use the SLIC~\cite{achanta2012slic} algorithm to generate $16$ superpixels on each frame. 
Following~\cite{henaff2021cont_det}, we alternatively use the graph-based image segmentation method~\cite{felzenszwalb2004fh_seg} to generate $16$ image segments for each frame. 
We use two scales to generate segments, the scale $s$ and minimum cluster size $c$, and $s=c \in \{500, 1000\}$ in practice.
%
After the segments generation, we convert each segment into its minimal bounding box and only keep those with width/height between $[0.05, 0.7]$ of the image width/height.

\noindent \textbf{Boxes from detectors.} 
We also use off-the-shelf modern detectors to generate object-centric bounding boxes for weakly supervised learning.
A CenterNet-based~\cite{zhou2019centernet} person detector is employed to generate bounding boxes on persons only. 
%
%
As an alternative, we use a generic object detector, which is based on Cascade RCNN~\cite{du2021crcnn}. 

\section{Downstream tasks}

\subsection{Action Recognition}
On all video action recognition datasets, we use the video clip of $32$ frames with temporal stride $2$ as input. 
During training, the temporally consistent random data augmentation~\cite{qian2021cvrl} of cropping, resizing and flipping are applied and the resolution is set to $224 \times 224$.
During evaluation, we densely sample $10$ clips with resolution $256 \times 256$ from each video and apply a 3-crop evaluation following~\cite{feichtenhofer2019slowfast}.
%

\noindent \textbf{Linear Evaluation.} 
On action recognition datasets, we train a linear classifier with fixed backbone weights using the SGD optimizer with momentum of $0.9$.
On Kineitcs400~\cite{kay2017k400}, the linear classifier is trained for $100$ epochs with learning rate of $32$ and batch size of $1024$.
On UCF101~\cite{soomro2012ucf101} and HMDB51~\cite{kuehne2011hmdb}, the linear classifier is trained for $50$ epochs with learning rate of $0.84$ and batch size of $128$.
No dropout and weight decay are applied.

\noindent \textbf{Fine-tuning.} 
On UCF101~\cite{soomro2012ucf101} and HMDB51~\cite{kuehne2011hmdb}, we use the pre-trained models to initialize the network and fine-tune all layers for $50$ epochs.
We use batch size of $128$, weight decay of $1\text{e-}5$ and dropout rate of $0.5$ during fine-tuning. 
The learning rate is set to $0.72$ and $0.84$ for UCF101 and HMDB51 respectively.

\subsection{Spatio-temporal Action Localization}
We use the same action transformer head as in~\cite{girdhar2019vat,li2020avak} to our R3D-50 backbone and follow the setting in~\cite{li2020avak}.
The model is fine-tuned with batch size $256$ for $50$k steps, which is around $36$ epochs on AVA-Kinetics~\cite{li2020avak}. 
The input has $32$ frames with resolution $400$ and temporal stride $2$.
We use the SGD optimizer with momentum $0.9$ during the fine-tuning.
On AVA-Kinetics, the learning rate is $1\text{e-}2$ and the weight decay is $1\text{e-}7$. 
On AVA~\cite{gu2018ava}, the learning rate is set to $3\text{e-}2$ and the weight decay is $1\text{e-}4$.
During evaluation, we use the same set of detected boxes in \cite{li2020avak} for AVA-Kinetics and in \cite{feichtenhofer2021rho_moco} for AVA v2.2 for a fair comparison.

\subsection{Object Tracking}
To evaluate on OTB2015~\cite{wu2013otb} dataset, we follow the same practice as in~\cite{xu2021vfs, gordon2020vince, yao2020seco} to adopt the SiameseFC~\cite{bertinetto2016siamese_fc} as the tracker.
Specifically, we modify the spatial stride and dilation rate to be $(1, 2)$ and $(1, 4)$ in the first layer of the $res_4$ and $res_5$ blocks.
These modifications allows us to increase the feature map resolution without impacting on the pre-trained model.
We fine-tune the tracker on GOT-10K~\cite{Huang2021got10k} dataset using the SGD optimizer with momentum of $0.9$.
We use batch size of $256$, learning rate of $0.1$ and weight decay of $1\text{e-}4$ and the tracker is fine-tuned for 20 epochs.
%

\section{Visualization}
\subsection{Attention}
\input{Sup_Materials/attention}
We visualize the learned attention map during the training in Figure~\ref{fig:attention}.
For visualization purpose only, we use the boxes from the object detector to pool the region features in the source views to generate the attention maps.
The model is trained with the randomly generated boxes as described in the paper.
In Figure~\ref{fig:attention}, we visualize the center frames in the source and the target views and the source frames are superimposed with one box for visualization.
The zoomed-in thumbnails are presented in the second column.
Given the context (features from the target views), we use these thumbnails' region feature as the query to generate the attention maps shown in the fourth column.
It is interesting to observe that the model learns to attend to not only the corresponding instance in the target view, but also to some other semantically meaningful objects the instance potentially interacts with.

\subsection{Visual Object Tracking}
\input{Sup_Materials/tracker}
We provide some qualitative results on visual object tracking on OTB2015~\cite{wu2013otb} in Figure~\ref{fig:tracker}.
The results show that our tracker could robustly track objects under different scenarios.

\end{appendix}

\end{document}

%% file: 1.Introduction.tex
\section{Introduction}
\label{sec:introduction}
Self-supervised learning (SSL) has revolutionized natural language processing~\cite{lan2020albert, devlin2018bert} and computer vision~\cite{chen2020simclr, grill2020byol, akbari2021vatt, alayrac2020mmv} due to strong representations learned from a vast amount of unlabeled data.
The key breakthroughs that paved the way for SSL's success in computer vision come from the instance discrimination pretext task~\cite{dosovitskiy2014discriminative} and the contrastive objective~\cite{oord2018info_nce}, with which for the first time the self-supervised pretraining surpasses the supervised pretraining on downstream visual tasks~\cite{he2019moco}.
For videos, many self-supervised contrastive learning approaches~\cite{qian2021cvrl,feichtenhofer2021rho_moco, akbari2021vatt, alayrac2020mmv} directly extend established image-based methods~\cite{he2019moco,chen2020simclr} to the spatio-temporal domain. 
Most of them, however, do not explicitly exploit the temporal evolutions of multiple instances and scene context in videos.

Self-supervised learning methods typically enforce semantic consistency across views to construct instance representations~\cite{he2019moco, chen2020simclr}. 
%
This assumption is particularly true in the image domain because two views are typically generated from the same image.
As shown in Fig.~\ref{fig:task_a}, the goal is to enforce the representations of these two views to be as close as possible in the feature space.
In the video domain, these view-based contrastive approaches~\cite{qian2021cvrl, feichtenhofer2021rho_moco} may be less effective as the visual appearance of an instance frequently and drastically changes across frames. 
For example, one person in a video can have different poses and perform different activities over time, indicating the states and semantics of an instance are likely to change across space and time.
Enforcing spatio-temporal persistency throughout the video~\cite{feichtenhofer2021rho_moco} would lead to representations only encoding minimally shared information across frames, which may negatively impact spatio-temporally fine-grained downstream tasks.

Furthermore, existing self-supervised methods typically focus on learning representations for holistic visual understanding tasks~\cite{chen2020simclr, qian2021cvrl}, such as image classification and video action recognition.
For dense prediction tasks, such as object detection, action localization and tracking, those models are enhanced by adding task specific heads.
On the other hand, several approaches are designed to learn discriminative local features for dense prediction tasks~\cite{wang2021densecl, yang2021inst_loc, gordon2020vince, xu2021vfs},
%
but their performances on holistic visual understandings are often compromised~\cite{xie2021detco}.
%
%
In light of this, we are interested in learning representations that can be applied to both holistic and local video tasks.
%

We propose Contextualized Spatio-Temporal Contrastive Learning (\ours), illustrated in Fig.~\ref{fig:task_b},
to circumvent the undesirable strong spatio-temporal persistency enforced by the global contrastive objective.
\ours learns semantically consistent but discriminative local representations for various video downstream tasks, 
ranging from spatio-temporal action localization and object tracking to action recognition.
Specifically, 
we design a projection function $g(\cdot, \cdot)$ to take not only the instance feature but also the context feature into account,
where the instance feature is extracted from the source view of a video, and the context feature is sampled from the target view.
%
%
This task enforces the model 
to be context-aware in a video and thus is a good proxy to learn discriminative local representations.
%

To address the imbalanced capability of learning holistic and local video representations, we design a simple two-branch module to facilitate the network to learn high-quality video representations both globally and locally in one unified self-supervised learning scheme.

%
We evaluate the learned representations on a variety of downstream tasks.
For holistic representations, we evaluate with video action recognition on Kinetics400~\cite{kay2017k400}, UCF101~\cite{soomro2012ucf101} and HMDB51~\cite{kuehne2011hmdb} datasets. 
For local representations, we conduct experiments with spatio-temporal action localization on AVA-Kinetics~\cite{li2020avak} and AVA~\cite{gu2018ava} datasets, and the single object tracking on the OTB2015~\cite{wu2013otb} dataset.
Our experimental results show that by pretraining with \ours, the learned representations 
adapt well across all studied datasets, surpassing recently proposed methods that use either supervised pretraining or the self-supervised pretraing~\cite{qian2021cvrl, xu2021vfs, feichtenhofer2021rho_moco, girdhar2019vat}.
%

The main contributions of this work are:
\begin{compactitem}
    \item A region-based contrastive learning framework for fine-grained spatio-temporal representation learning.
    \item A contextualized region prediction task that facilitates learning semantically consistent while locally discrinimative video features.
    \item A simple network design to effectively reconcile simultaneous holistic and local representation learning.
    \item Competitive performance on 6 benchmarks, including spatio-temporal action localization, object tracking, and video action recognition.  
    %
    
\end{compactitem}

%% file: 2.RelatedWork.tex
\section{Related Work}
\label{sec:related_work}

\noindent \textbf{Self-supervised learning in images.}
To effectively learn representations from images, early self-supervised methods focus on designing pretext tasks by experts.
Various pretext tasks have been proposed, including colorization~\cite{larsson2017colorization}, inpainting~\cite{pathak2016context}, denoising~\cite{vincent2008extracting}, egomotion prediction~\cite{agrawal2015learning}, context prediction~\cite{doersch2015unsupervised}, orientation prediction~\cite{gidaris2018rotations}, spatial jigsaw puzzle~\cite{noroozi2016jigsaw}, \etc. 
Recent advances in image self-supervised learning stem mainly from minimizing contrastive loss~\cite{oord2018info_nce} on instance discrimination tasks~\cite{dosovitskiy2014discriminative}.
The contrastive objective effectively enforces representations of the same instance from different views to be similar, while it repels representations from different instances in the latent space. 
Representative frameworks in this category include NPID~\cite{wu2018npid}, MoCo~\cite{he2019moco, chen2020moco_v2}, SimCLR~\cite{chen2020simclr}, \etc.

\vspace{1mm}
\noindent \textbf{Self-supervised learning in videos.} 
In the video domain, self-supervised representation learning prospers in recent years.
%
%
Contrastive objectives have been widely used to learn video representations for holistic recognition tasks~\cite{qian2021cvrl, feichtenhofer2021rho_moco, recasens2021brave, yang2020tempo, singh2021semi}.
Extensive pretext tasks have been exploited to learn good representation in videos. 
Compared with the image domain, videos naturally yield richer self-supervision signals. 
In~\cite{recasens2021brave}, the goal is to enforce global context consistency and utilize long-short views of a video to align representations.
In~\cite{wang2018pulling}, the action and context features are factorized separately by learning from the conjugate examples in video dataset.
Motion signals are also exploited for learning good representations~\cite{tian2020contrastive, han2020self-motion}.
Temporal ordering of frames in a video has also been used for self-supervised representation learning.
%
For instance, in~\cite{misra2016shuffle, lee2017sorting}, temporal ordering of frames is enforced to learn representations by scuffling frames in a self-supervised manner.  
Similarly, forward and backward ordering of frames is used as the self-supervision signals for representation learning~\cite{wei2018arrow}.
In addition, temporal cycle consistency is exploited to learn spatio-temporal correspondence between video frames~\cite{jabri2020random_walk,wang2019cycle}.
On the other hand, multi-modal signals, such as audio/visual and visual/text, have been used to learn representations in a self-supervised manner that outperform models based on a single modality~\cite{alayrac2020mmv, alwassel2019self, patrick2021compositions, morgado2020audio}.  

\vspace{1mm}
\noindent \textbf{Local representations.} 
Although existing methods focus on learning holistic representations on images or videos, several recent approaches explicitly model spatially fine-grained representations.
In~\cite{chaitanya2020contrastive_global_local, wang2021densecl},
several constriave learning models have been developed for dense prediction tasks, such as object detection and image segmentation.
In addition, augmented samples with 
pseudo ground-truth are generated to learn dense features for object detection~\cite{yang2021inst_loc, ding2021patch_reid}.
Other methods introduce location priors to group pixels for learning local features. For example,~\cite{zhang2020self_xiao, henaff2021cont_det, van2021unsupervised_van_gool} use unsupervised masks and~\cite{pinheiro2020vader, xie2021propagate} use pixel coordinates.
In the video domain, 
many methods learn fine-grained features by leveraging inherent temporal augmentations to determine object correspondence~\cite{xu2021vfs, yao2020seco, gordon2020vince, jabri2020random_walk}.
\cite{gordon2020vince} randomly samples two images from a video to construct the contrastive pairs for self-supervised learning and shows improved performance on video tasks.
\cite{yao2020seco} employs pretext tasks of determining whether frames are from the same video and their temporal ordering.
\cite{xu2021vfs} observes the emergence of correspondence by learning frame-level similarity.
\cite{jabri2020random_walk} enforces forward-backward temporal consistency to learn local correspondences in videos.
%
Most of the existing approaches focus on learning local representations and do not emphasize performances on holistic tasks. \cite{xie2021detco} explicitly raises the question about simultaneously learning holistic and local representations but only focuses on the image domain.

Different from these related works, our method leverages video context during self-supervised learning by employing a novel region-based prediction tasks, and is designed to learn holistic and local representations simultaneously with self-supervision from unlabeled videos. 
Unlike most of the aforementioned methods that focus on learning either holistic or local representations, we emphasize the quality of both under one unified training scheme.

%% file: 3.Method.tex
\section{Method}
\label{sec:method}

In this section, we introduce the proposed self-supervised learning framework, Contextualized Spatio-Temporal Contrastive Learning (\ours) for learning spatio-temporally fine-grained representations in videos.

\input{Tex/Method/1.instance_constrastive_learning}

\begin{figure}
    \centering
    \includegraphics[width=\columnwidth]{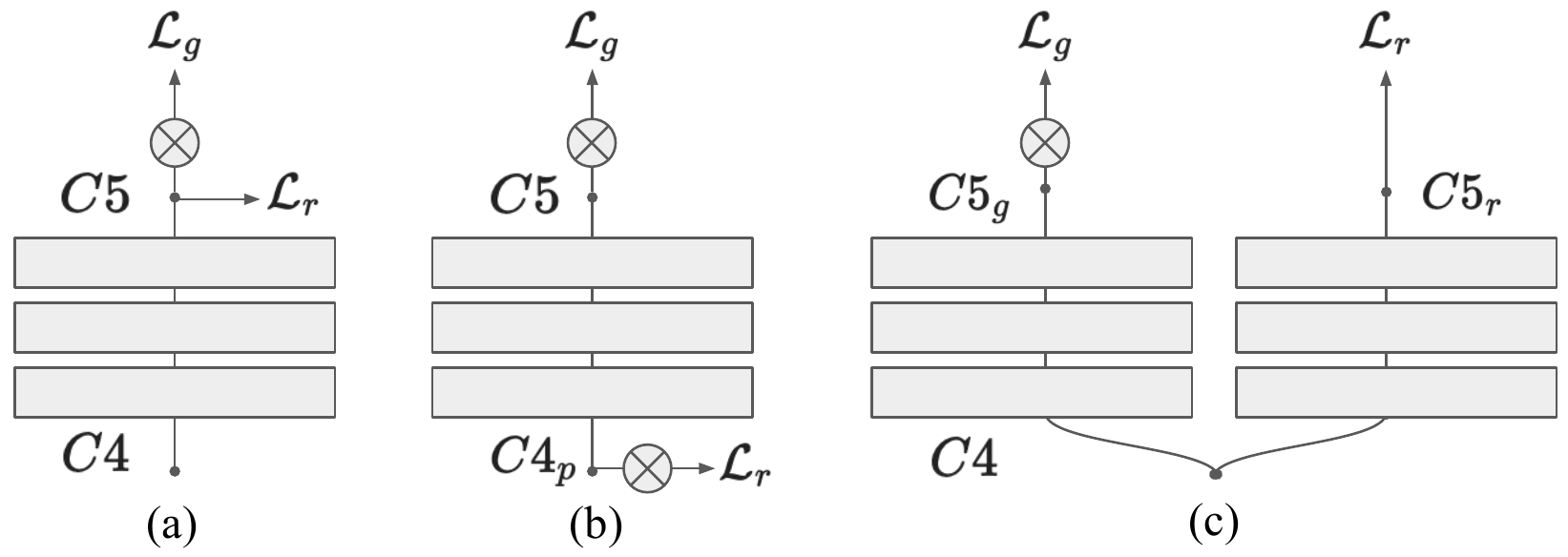}
    \vspace{-15pt}
    \caption{
    \textbf{Balancing global and local losses.}
    We analyze three different endpoints to impose global and local training losses. 
    $\otimes$ indicates a pooling operation.
    Throughout the experiments, we find that by branching the res5 block in the network and applying $\mathcal{L}_g$ and $\mathcal{L}_r$ on R$5_g$ and R$5_r$ respectively, two losses mutually benefit and the representations thrive by co-training.
    }
    \label{fig:global_local_loss}
    \vspace{-5pt}
\end{figure}

\input{Tex/Method/2.contextualized_instance_prediction}
\input{Tex/Method/3.instance_locations.tex}

\input{Tables/main_table.tex}

\input{Tex/Method/4.balance_global_local.tex}

%% file: Tex/Method/1.instance_constrastive_learning.tex
\subsection{Region-Based Contrastive Learning in Videos}
\label{sec:icl}
%
%
Given a video, a simple contrastive learning algorithm randomly samples two video clips $\{x, x^\prime\}$, and applies random data augmentation on each video clip independently.
Corresponding video-level representations $\{z, z^{\prime}\}\in\mathbb{R}^C$ are extracted by the network $f(\cdot)$ for computing a contrastive loss~\cite{oord2018info_nce}, with negative examples from other videos.
We denote this video-level global contrastive loss as $\mathcal{L}_g$.
This training objective enforces globally average-pooled features from the same video to be similar while it repels such features from different videos.
However, no explicit supervision is enforced on local features, which play an important role for dense prediction tasks.

%
To enforce local supervision, one way is to extend~\cite{wang2021densecl} to the spatio-temporal domain.
Given the dense feature maps $\{F, F^{\prime}\}\in\mathbb{R}^{T \times H \times W \times C}$ from $\{x,x^{\prime}\}$, where $T, W, H, C$ are time, height, width and channel dimention respectively, for each feature voxel $h_i \in F$, we find its correspondence $h_j^{\prime} \in F^{\prime}$ that it is closest to in the feature space to form a positive pair.
%
%
Thus, the dense contrastive loss in a video can be formulated as:
\begin{align}
    z_i &= g(h_i) = \text{MLP}(h_i),\label{eq:dense_cl_z} \\
    \mathcal{L}_{r} &= \sum_i -\log \frac{\exp(z_i \cdot z_j^{\prime}/\tau)}{\exp(z_i \cdot z_j^{\prime}/\tau) + \sum_k {\exp(z_i \cdot \hat{z}_k/\tau)}}, \notag \\
    \mbox{s.t.} &\quad j = \argmin_j h_i \cdot h^\prime_j,\label{eq:dense_cl_loss}
\end{align}
where MLP refers to a multi-layer perceptron, $\tau$ is the temperature parameter, $i$, $j$ and $k$ are grid indices, and $\{\hat{z}\}$ are representations from other videos.
Here, we simply regard all the dense features from other videos as negative examples for loss computation. 

Assuming that we have access to the location priors of regions of interest $\{r_i\}$, we can derive the vanilla region-based contrastive learning by organizing the representations using the region location:
\begin{align}
    z_i &= g(h_i) = \text{MLP}(\text{ROIAlign}(F, r_i)),\label{eq:unary_region_z}
\end{align}
where we override the notation $h_i$ to be the pooled region features, with $i$ being the region index. 
In this paper, we parameterize a region as a bounding box of a certain frame $r = \{t, x_{\textrm{min}}, y_{\textrm{min}}, x_{\textrm{max}}, y_{\textrm{max}}\}$. 
The region representation $h_i$ is pooled from the dense feature map $F$ by ROIAlign~\cite{ren2015faster-rcnn}. 

The full learning objective is the linear combination of the global loss and the local loss weighted by the scale factor $\omega$. And we average over the $N$ mini-batch during training: 
\begin{align}
    \mathcal{L} = \frac{1}{N} \sum \left(\mathcal{L}_g + \omega \mathcal{L}_r\right).
\end{align}

%% file: Tex/Method/2.contextualized_instance_prediction.tex
\subsection{Contextualized Spatio-Temporal Contrastive Learning (\ours)}
\label{sec:contextualization}
The vanilla region-based contrastive learning framework described in Section~\ref{sec:icl} has one limitation: the loss always encourages the representations of the same instance at different timestamps to be similar, 
while the appearance of an instance in a video may change across frames.
%
For example, one person in a video can appear in different poses and perform different activities.
Simply enforcing the similarity of the same instance at different temporal locations of the video will inadvertently encourage the model to encode only the minimal information, which is less effective for downstream video understanding tasks. 

To resolve this issue, we propose a novel self-supervised method, Contextualized Spatio-Temporal Contrastive Learning (\ours).
In a nutshell, \ours requires the network to learn to ``reconstruct" the representation of a region in a target view given its representation from the source view and the context features around the target view.
\begin{align}
    z_i &= g(h_i, F_c^\prime) = g(\text{ROIAlign}(F, r_i), F_c^\prime),\label{eq:binary_region_z} \\
    \mathcal{L}_r &= \sum_i -\log \frac{\exp(z_i \cdot h_j^{\prime}/\tau)}{\exp(z_i \cdot h_j^{\prime}/\tau) + \sum_k {\exp(z_i \cdot \hat{h}_k/\tau)}}, \notag \\
    \mbox{s.t.} &\quad j = \argmin_j h_i \cdot h^\prime_j,\label{eq:inst_loss}
\end{align}
where $F^{\prime}_c$ denotes the set of context features around the target view, and $i$, $j$ and $k$ are region indices, and $\{\hat{h}\}$ are representations from other videos.
Here we would like to note: (1) comparing with Eq.~(\ref{eq:unary_region_z}), we extend the representation decoding function $g(\cdot)$ from an unary function to a binary function $g(\cdot,\cdot)$ in Eq.~(\ref{eq:binary_region_z});
(2) we do not force bijective mapping between regions in two clips, so the different numbers of regions between views do not cause a problem.
Eq.(\ref{eq:binary_region_z}-~\ref{eq:inst_loss}) formulate ConST-CL in a general manner, which considers all regions from all frames. 
It might pose computation challenges, so in practice, we instead construct two sets of regions by sampling from one temporal slice of each feature map. 
The temporal sampling strategy will be discussed in the following ablation studies.

Fig.~\ref{fig:method} shows an illustration of \ours. 
Given a pooled region feature $h_i$ from the source view and a set of context features $F^\prime_c$ from the target view, a projection function $z_i = g(h_i, F^\prime_c)$ is learned with the objective to minimize the representation distance between the reconstructed relating representation $z_i$ and its corresponding native representation $h_j^\prime$ in the target view.
The context features $F^\prime_c$ are a subset of feature voxels sampled from the dense feature map $F^\prime$.
In our case, we subsample a few frames from the dense video representations $F^\prime$ along the temporal dimension.
We define the number of frames used to construct $F^\prime_c$ as the \textit{context length}, whose effect on the performance is studied in Section~\ref{sec:ablations}.
Different from simply contrasting features that are projected into the shared feature space, \ours requires every feature vector in $\{F, F^\prime\}$ to encode more information about itself and the context, such that with context features from another view, $g(\cdot, \cdot)$ can reconstruct the instance encoding conditionally.

We implement $g(\cdot,\cdot)$ using a transformer~\cite{vaswani2017attention} architecture.
First, we linearly project each instance feature vector from the source view $h_i$ to a Query token, and the context feature vectors from the target view $F_c^\prime$ to Key-Value token pairs. 
The multi-head cross-attention is then used to look up the Key-Value pairs by the Query token.
Finally, we apply the InfoNCE loss~\cite{oord2018info_nce} on the transformed instance feature $z_i$ and its correspondence $h_j^\prime$ using Eq.~(\ref{eq:inst_loss}).

%% file: Tex/Method/3.instance_locations.tex
\subsection{Region Generation}
\label{sec:box_generation}

\noindent \textbf{Random boxes.} Generating random boxes on-the-fly during the training is the most straightforward method. 
For all of our related experiments, we randomly generate $8$ boxes on each frame. 
%
%
In Section~\ref{sec:ablations}, we show that our method performs interestingly well trained with these random boxes.

\noindent \textbf{Boxes from low-level image cues.}
We also consider two methods to generate boxes from low-level image cues. 
Specifically, we use the SLIC~\cite{achanta2012slic} algorithm to generate $16$ superpixels on each frame. 
Following~\cite{henaff2021cont_det}, we alternatively use the graph-based image segmentation method~\cite{felzenszwalb2004fh_seg} to generate $16$ image segments for each frame. 
%
%

\noindent \textbf{Boxes from detectors.} 
We also use off-the-shelf modern detectors to generate object-centric bounding boxes for weakly supervised learning.
A CenterNet-based~\cite{zhou2019centernet} person detector is employed to generate bounding boxes on persons only. 
%
%
As an alternative, we use a generic object detector, which is based on Cascade RCNN~\cite{du2021crcnn}. 


%% file: Tables/main_table.tex
\begin{table*}[t!]
\centering
\small
\begin{tabular}{ccccccccccccccc} 
\hline
 
 &  & Pre-training & \multicolumn{2}{c}{AVA-Kinetics} & & \multicolumn{2}{c}{AVA} & & \multicolumn{2}{c}{Object Tracking} \\
\cline{4-5} \cline{7-8} \cline{10-11}
Method & Backbone & Dataset & mAP (GT) & mAP (Det) & & mAP (GT) & mAP (Det) & & Precision & Success \\

\hline
INet-sup~\cite{li2020avak} & I3D & K400 & $35.9$ & $22.9$ & & 27.5 & 19.1 & & - & -\\
K400-sup & R3D50 & K400 & $26.7$ & $19.8$ & & - & 22.2 & & $71.2$ & $51.46$ \\
\hline
SimSiam~\cite{chen2021simsiam} & R50 & INet & - & - & & - & - & & $61.0$ &  $43.2$ \\
VINCE~\cite{gordon2020vince} & R50 & R2V2 & - & - & & - & - & & $66.0$ & $47.6$ \\
SeCo~\cite{yao2020seco} & R50 & K400 & - & - & & - & - & & $71.9$ &  $51.8$ \\
VFS~\cite{xu2021vfs} & R50 & K400 & - & - & & - & - & & $73.9$ &  $52.5$ \\
\hline
$\rho$MoCo~\cite{feichtenhofer2021rho_moco} & Slow-only & K400 & - & - & & - & $20.3$ & & - & - \\
$\rho$BYOL~\cite{feichtenhofer2021rho_moco} & Slow-only & K400 & - & - & & - & $23.4$ & & - & - \\
VFS-inflated & R3D50 & K400 & $34.6$ & $25.9$ & & $29.1$ & $22.4$ & &  $73.3$ & $52.7$ \\
CVRL~\cite{qian2021cvrl} & R3D50 & K400 & $31.6$ & $24.1$ & & $24.9$ & $18.4$ & & $75.4$ & $53.7$ \\
\hline
\ours & R3D50 & K400 & $\mathbf{39.4}$ & $\mathbf{30.5}$ & & $\mathbf{31.1}$ & $\mathbf{24.1}$ & & $\mathbf{78.1}$ & $\mathbf{55.2}$ & \\

\hline
\end{tabular}
\vspace{-5pt}
\caption{\textbf{Downstream task performances based on pre-trained representations.} The learned representations are evaluated for spatio-temporally fine-grained tasks, including spatio-temporal action recognition on the AVA v2.2 and AVA-Kinetics (using both ground-truth and detected person boxes) and single object tracking on OTB2015.
\ours achieves the state-of-the-art results across the board, suggesting the effectiveness of our proposed framework that is capable of coherently learning better local visual representations in videos.}
\label{tab:main_table}
\vspace{-5pt}
\end{table*}

%% file: Tex/Method/4.balance_global_local.tex
\subsection{Balancing Global and Local Losses}
Existing methods~\cite{zhou2016cams, xu2021vfs} have shown discriminative local features can be extracted 
by applying supervisory signals on holistic representations. 
Intuitively, adding constraints on both holistic and local representations are mutually beneficial, because discriminative local features would contribute to holistic recognition, while expressive holistic features could be derived from local features.
In practice, however, we find that directly adding the proposed region-based local loss on the dense feature map right before the average pooling layer in the ResNet, the self-supervised training is less stable and sensitive to the hyper-parameters for balancing the global and local losses. 

To address this issue, we propose a simple solution. 
As we use ResNet3D-50 as our base model, instead of adding the local loss on the $C5$ endpoint, we modify the ResNet architecture and replicate the $res5$ block, forming a ``Y'' structure, as shown in Fig.~\ref{fig:global_local_loss}.
Then the global and local losses are attached on endpoint $C5_g$ and $C5_r$, respectively, and they co-constrain the latent feature map in $C4$ during training.
When fine-tuning the model for downstream tasks, we take either $C5_g$ or $C5_r$ branch depending on the task.
This design introduces only moderate additional computes during the pre-training stage and is at no extra cost for fine-tuning and inference. 
We will show in Section~\ref{sec:ablations} that the proposed ``Y'' structure results in better trade-off for both video-level and instance-level downstream tasks.

%% file: 4.Experiments.tex
\section{Experiments}
\label{sec:experiments}

%
We evaluate clip-level video representation models on the Kinetics400~\cite{kay2017k400} dataset following the linear probing protocol, and the UCF101~\cite{soomro2012ucf101} and  HMDB51~\cite{kuehne2011hmdb} datasets using both linear probing and fine-tuning.
To evaluate the learned spatio-temporally fine-grained representations, we conduct experiments on the  AVA-Kinetics~\cite{li2020avak} and AVA v2.2~\cite{gu2018ava} datasets for spatio-temporal action localization and the OTB2015~\cite{wu2013otb} dataset for single object tracking.

\input{Tex/Experiments/1.experiment_setup.tex}
\input{Tex/Experiments/2.downstream_evaluations}
\input{Tex/Experiments/3.main_results.tex}
\input{Tables/video_classification.tex}
\input{Tables/model_size.tex}

\input{Tables/temporal_sampling.tex}
\input{Tables/endpoints.tex}

\input{Tables/context_length_and_overhead.tex}
\input{Tables/box_type.tex}
\input{Tex/Experiments/4.ablation_studies}

\subsection{Limitations}
One missing piece in the current framework is the self-supervisory signal for learning even finer-grained representations.
We hope to enrich our method by incorporating dense self-supervision in the future. 
Moreover, the current solution of organizing global-local self-supervisory signals is limited to the convolutional neural network backbone.
For the recent vision transformer (ViT)~\cite{dosovitskiy2020vit}, it is non-trivial to apply our proposed method directly in its current form. 

%% file: Tex/Experiments/1.experiment_setup.tex
\subsection{Implementation Details}
\label{sec:implementation_details}
We use the ResNet3D-50 (R3D50) as our backbone feature extractor following~\cite{qian2021cvrl}.
All features are $\ell_2$ normalized before being used to compute the self-supervised loss.

For the holistic representation learning branch, we use a 3-layer MLP with $2048$ hidden nodes to project a $2048$-dimensional feature vector into a $128$-dimensional feature vector.
On the local representation learning branch, we use the same attention-based architecture as described in~\cite{vaswani2017attention}. 
The attention units are stacked into multiple heads and layers to construct the \ours head for the instance prediction task.
In this work, the head of \ours consists of 3-layer 3-head attention units with hidden dimension of $128$.
We use the ReLU activation function without dropout. 
A final linear layer is used to project the $128$-dim feature vector back to $2048$-dim.
We add spatio-temporal positional encodings to the query, key and value tokens before inputting into the transformer head in order to preserve location information. 
To construct contrastive pairs for local branch, we always sample examples from the center frame in both views for the experiments unless otherwise specified. 
The self-supervised pre-training is performed on the Kinetics400~\cite{kay2017k400} dataset.
During evaluations, all the heads used for self-supervised learning are discarded.

All models are pre-trained with mini-batch of $1024$. 
During the pre-training, we use the SGD optimizer with momentum of $0.9$.
The learning rate is linearly warmed-up to $40.96$ during the first $5$ epochs, followed by half-period cosine learning rate decay~\cite{he2019bag_of_tricks} to $0$. 
A weight decay of $10^{-6}$ is applied to all kernels.
We set the temperature $\tau$ to be $0.1$ for the global loss and $0.2$ for the local loss. 
The scale factor $\omega$ is $0.01$ to balance the global and local losses.

For results in Table (\ref{tab:main_table},\ref{tab:classification_table}), we pre-train the backbone model for $200$k steps, which is around $850$ epochs on the Kinetics400 dataset with the randomly generated region boxes. 
And the context length is set to $5$.
For all ablation studies, we use backbone models from a shorter pre-training schedule, which trains for $100$k steps.
%

%% file: Tex/Experiments/2.downstream_evaluations.tex
\subsection{Downstream Tasks}
%
It is of great interest to understand whether one representation model can be applied to both holistic and local understandings, as intuitively the better local representations can facilitate holistic recognition tasks and vice versa.
In this work, we apply the learned representation models to (1) video action recognition tasks that require holistic representations on the  Kinetics400~\cite{kay2017k400}, UCF101~\cite{soomro2012ucf101} and HMDB51~\cite{kuehne2011hmdb} datastes; and (2) 
spatio-temporal action localization and single object tracking  tasks that require local representations on the  AVA-Kinetics~\cite{li2020avak}, AVA v2.2~\cite{gu2018ava} and OTB2015~\cite{wu2013otb} datasets. 

\noindent \textbf{Video action recognition.}
we perform linear evaluation by fixing all the backbone weights on Kinetics400~\cite{kay2017k400}, UCF101~\cite{soomro2012ucf101} and HMDB51~\cite{kuehne2011hmdb}. 
The input is a $32$-frame video clip with temporal stride of $2$ and resolution of $224$. We train the linear classifier for $100$ epochs.
%
%
We also use the pre-trained models to initialize the network and fine-tune all layers for $50$ epochs on UCF101 and HMDB51.

\noindent \textbf{Spatio-temporal action localization.} We attach the same action transformer head as in~\cite{girdhar2019vat,li2020avak} to our R3D50 backbone, and follow the setting in~\cite{li2020avak} for simplicity.
We train our model with ground-truth person bounding boxes and use either ground-truth boxes or boxes generated using off-the-shelf person detectors\footnote{We use the same set of boxes as in~\cite{li2020avak, feichtenhofer2021rho_moco} for fair comparisons.} for region proposals during evaluation. 
%
The model is trained with batch size $256$ for $50$k steps.
%
The input has $32$ frames with resolution of $400$ and temporal stride of $2$.

\noindent \textbf{Single object tracking.} 
We also evaluate our learned representations via single object tracking task, which requires semantically consistent spatio-temporal features to determine object-level correspondence.
We follow the same practice as in~\cite{xu2021vfs, gordon2020vince, yao2020seco} to adopt the SiameseFC~\cite{bertinetto2016siamese_fc} as the tracker and modify the spatial stride and dilation rate in the $res_4$ and $res_5$ blocks. 
Note that our backbone is a 3D-convolutional network, and for each input frame we also sample its neighboring $n$ frames from each side and use the $2n+1$ frames as the input. 
After the $res_5$ block, we slice the center frame along the time dimension of the local feature map $F$ for the input to the tracking head.
Here we use $n=2$ as the largest temporal kernel in the network is $5$.
We use the pre-trained checkpoint to initialize the backbone and fine-tune the tracker for all experiments.

%% file: Tex/Experiments/3.main_results.tex
\subsection{Main Results}

In Table~\ref{tab:main_table}, we study the model performance on dense vision tasks by using the pre-trained models from different methods.
We evaluate the spatio-temporal action localization on the AVA-Kinetics~\cite{li2020avak} and AVA v2.2~\cite{gu2018ava}.
Following~\cite{li2020avak, girdhar2019vat}, the models are evaluated under two settings: using either ground-truth boxes or detected boxes as region proposals on the validation set.
On AVA-Kinetics, \ours achieves $39.4\%$ mAP when using ground-truth boxes and $30.5\%$ mAP when using detected boxes, outperforming the supervised method~\cite{li2020avak} by a large margin.
\ours also outperforms the baseline self-supervised method CVRL~\cite{qian2021cvrl} model 
with more than $24\%$ relative performance gain. 
In addition, we compare to the VFS~\cite{xu2021vfs} which is designed for dense vision tasks. 
As VFS uses 2D ResNet, we follow the common practice to inflate all 2D kernels in the network into 3D~\cite{carreira2017i3d} and load the pre-trained weights from VFS for the fair comparison.
In the table, \ours outperforms the VFS-inflated method by more than $4.6\%$ mAP, showing the effectiveness of our proposed method on spatio-temporal action recognition task.
On the AVA v2.2, we observe similar trend that \ours outperforms competing methods, achieving $31.1\%$ and $24.1\%$ mAP using ground-truth and detected boxes respectively.


On OTB2015~\cite{wu2013otb}, we first compare with prior methods designed for dense task \textit{only}.
Table~\ref{tab:main_table} shows that 
\ours outperforms the evaluated methods by a large margin.
Specifically, compared to VFS~\cite{xu2021vfs}, \ours achieves $78.1\%$($+\Delta 4.2\%$) in precision score and $55.2\%$($+\Delta 2.7\%$) in success score.
To rule out the effect of architecture difference (2D network vs.\ 3D network), we inflate the 2D ResNet into 3D and load the VFS pre-trained checkpoint, denoted as VFS-inflated in the table.
Compared to VFS, VFS-inflated performs similarly to its 2D counterpart, which indicates the effect of this architecture difference on the tracking task is insignificant.
%
%
When compared with CVRL, \ours achieves clear performance gain for single object tracking on the OTB2015 benchmark. 
%

In Table~\ref{tab:classification_table}, our method performs comparable to CVRL and $\rho$MOCO ($\rho$=2) using linear probing and achieves competitive fine-tuning results with top-1 accuracy of $94.8\%$ and $71.9\%$ on UCF101 and HMDB51, respectively. 
It is worth pointing out that our method improves upon CVRL on fine-tuning UCF101 and HMDB51, even though we do not use any extra supervision on holistic representations other than the CVRL's loss $\mathcal{L}_g$. 
These findings are consistent with our intuition that the holistic and the local representation modeling can be mutually beneficial. 
%
In our method, the two losses simultaneously contribute to and constrain on the latent feature map $C4$ in the network. 
These results also demonstrate the effectiveness of the proposed model design that coherently organizes different levels of representations in a single framework.
%


Finally, in Table~\ref{tab:model_size} we report the model size and computational cost including \textit{both} the backbone and the SSL heads in different video self-supervised learning methods. 
Comparing to CVRL, \ours increased model sizes and computational cost moderately, mainly due to the branched $res_5$ block and additional transformer head.
The Slow-only network and training strategy used in~\cite{feichtenhofer2021rho_moco} is different from ours, making the side-to-side comparison difficult. 
Thus we leave the results in the table for reference. 

%% file: Tables/video_classification.tex
\begin{table}[t!]
\centering
\small
\setlength{\tabcolsep}{.6\tabcolsep}
\begin{tabular}{ccccccccccc} 
\hline
 & \multicolumn{3}{c}{Linear} & & \multicolumn{2}{c}{Fine-tune} \\
\cline{2-4} \cline{6-7} \cline{9-10}
Method & K400 & UCF & HMDB & & UCF & HMDB \\

\hline
VINCE~\cite{gordon2020vince} & $49.1$ & - & - & & - & - \\
SeCo~\cite{yao2020seco} & $61.9$ & - & - & & $88.3$ & $55.6$ & \\
\hline
VFS-inflated~\cite{xu2021vfs} & $33.1$ & - & - & & $71.4$ & $41.0$ \\
$\rho$MoCo ($\rho$=$2$)~\cite{feichtenhofer2021rho_moco} & $67.4$ & - & - & & $93.2$ & - \\
$\rho$BYOL ($\rho$=$4$)~\cite{feichtenhofer2021rho_moco} & $\mathbf{71.5}$ & - & - & & $\mathbf{95.5}$ & $\mathbf{73.6}$ \\
CVRL~\cite{qian2021cvrl} & $66.1$ & $\mathbf{89.2}$ & $57.3$ & & $92.2$ & $66.7$ \\
\hline
\ours & $66.6$ & $89.1$ & $\mathbf{59.9}$ & & $94.8$ & $71.9$ \\
\hline
\end{tabular}
\caption{\textbf{Downstream video action recognition.} 
%
\ours achieves competitive results on fine-tuning experiments, indicating it can learn strong holistic visual representations in videos.
}
\label{tab:classification_table}
\vspace{-5pt}
\end{table}

%% file: Tables/model_size.tex
\begin{table}[t!]
\centering
\small
\begin{tabular}{cccc} 

\hline
Method & Frames & Params (M) & FLOPs (G) \\
\hline
CVRL~\cite{qian2021cvrl} & 16$\times$2 & $44.6$ & $91.2$ \\
$\rho$MOCO ($\rho$=$2$)~\cite{feichtenhofer2021rho_moco} & 8$\times$2 & $44.6$ & $83.6$ \\
$\rho$MOCO ($\rho$=$2$)~\cite{feichtenhofer2021rho_moco} & 16$\times$2 & $44.6$ & $167.0$ \\
$\rho$BYOL ($\rho$=$4$)~\cite{feichtenhofer2021rho_moco} & 16$\times$4 & $44.6$ & $334.0$ \\
\ours & 16$\times$2 & $71.7$ & $113.0$ \\
\hline

\end{tabular}
\vspace{-5pt}
\caption{\textbf{Model sizes and computational costs} for different SSL methods with video-based networks. The SSL heads are included for the parameter and FLOPS counts.}
\label{tab:model_size}
\vspace{-5pt}
\end{table}

%% file: Tables/temporal_sampling.tex
\begin{table}[t!]
\centering
\small
\begin{tabular}{cccccc} 
\hline
 
Method & Sampling & UCF & HMDB & AVA-K & OTB \\

\hline
CVRL & - & $91.6$ & $66.2$ & $30.9$ & $75.9$ \\
\ours & Random & $56.2$ & $57.8$ & $34.8$ & $75.4$ \\
\ours & Center & $94.1$ & $67.7$ & $36.9$ & $77.1$ \\
\ours & Nearest & $93.8$ & $68.1$ & $36.9$ & $76.4$ \\
\hline

\end{tabular}
\vspace{-5pt}
\caption{
\textbf{Ablation on the temporal sampling strategy.}
The ``Center" and ``Nearest" temporal sampling strategy perform equally well and better than ``Random" sampling for \ours.}
\label{tab:sampling}
\vspace{-5pt}
\end{table}

%% file: Tables/endpoints.tex
\begin{table}[t!]
\small
\centering
\begin{tabular}{cccccc} 
\hline
 
Method & Endpoint & UCF & HMDB & AVA-K & OTB \\

\hline
CVRL & - & $91.6$ & $66.2$ & $30.9$ & $75.9$ \\
\ours & $C4_p$ & $93.5$ & $67.5$ & $32.0$ & $75.3$ \\
\ours & $C5$ & $93.4$ & $66.7$ & $33.6$ & $74.3$ \\
\ours & $C5_g$+$C5_r$ & $94.3$ & $68.7$ & $36.7$ & $77.7$ \\
\hline
\end{tabular}
\vspace{-5pt}
\caption{
\textbf{Ablation on the loss endpoints.}
We apply region-based contrastive loss on different endpoints and show that the $C5_g$+$C5_r$ configuration achieves the best trade-off between the global and local losses with the best downstream task performances.}
\label{tab:endpoints}
\end{table}

%% file: Tables/context_length_and_overhead.tex
\begin{table*}[t!]
\centering
\small
\begin{tabular}{ccccccccccc} 

\hline
Context & Loss & UCF & HMDB & \multicolumn{2}{c}{AVA-Kinetics} & &  \multicolumn{2}{c}{OTB}\\
\cline{5-6} \cline{8-9}
length & endpoint(s) & Top-1 & Top-1 & mAP (GT) & mAP (Det) & & Precision & Success & $\#$Params (M) & $\#$FLOPs (G) \\
\hline
- & R5  & $91.6$ & $66.2$ & $30.9$ & $23.4$ & & $75.9$ & $53.6$ & $44.6$ & $45.6$ \\
$0$ & 2R5 & $91.8$ & $66.0$ & $35.3$ & $27.5$ & & $75.4$ & $56.6$ & $77.7$ & $55.5$ \\
$1$ & 2R5 & $93.4$ & $66.7$ & $36.7$ & $28.0$ & & $77.7$ & $55.0$ & $71.7$ & $55.6$ \\
$3$ & 2R5 & $93.7$ & $67.4$ & $36.9$ & $28.1$ & & $77.5$ & $54.5$ & $71.7$ & $56.1$ \\
$5$ & 2R5 & $93.4$ & $67.5$ & $37.6$ & $28.1$ & & $79.0$ & $55.4$ & $71.7$ & $56.5$ \\
\hline

\end{tabular}
\caption{
\textbf{Ablation study on the context length and the computational overhead.}
``-" indicates no \ours is used and the model is only trained with the $\mathcal{L}_g$; 
``0" indicates no context is provided and the model simply degrades to the vanilla region-based contrastive learning.
From the table, we observe the trend that more context is helpful to learn better spatio-temporal representations.
We notice the number of parameter increment is largely from the duplicated $res5$ block while our transformer based head is more parameter efficient than the MLP head for the region-based contrastive learning.
}
\vspace{-5pt}
\label{tab:context_length_and_overhead}
\end{table*}

%% file: Tables/box_type.tex
\begin{table}[t!]
\centering
\small
\begin{tabular}{cccccc} 

\hline
 & UCF & AVA-K & \multicolumn{2}{c}{OTB} \\
\cline{4-5}
Box type & Top-1 & mAP (GT) & Precision & Success \\
\hline
Random & $94.3$ & $36.9$ & $77.0$ & $54.0$ \\
SLIC~\cite{achanta2012slic} & $93.4$ & $36.4$ & $76.3$ & $53.7$ \\
FH~\cite{felzenszwalb2004fh_seg} & $94.1$ & $36.9$ & $77.1$ & $54.3$ \\
Person Detector & $93.4$ & $36.7$ & $77.7$ & $55.0$ \\
Object Detector & $93.7$ & $37.2$ & $77.8$ & $54.1$ \\
\hline

\end{tabular}
\caption{
\textbf{Ablation study on the box type.}
When applying different types of boxes during the self-supervised training, 
we find that model learns equally well regardless of the region location quality.}
\label{tab:box_type}
\vspace{-5pt}
\end{table}

%% file: Tex/Experiments/4.ablation_studies.tex
\subsection{Ablation Study}

\label{sec:ablations}

\noindent \textbf{Temporal sampling strategy.}
%
To construct the contrastive region pairs, we sample one frame from the source and the target clip respectively.
In Table~\ref{tab:sampling}, we study three different temporal sampling strategies.
For ``random" strategy, we randomly sample frames from the source and the target views to construct the contrastive pairs.
For ``center" strategy, we simply choose the center frame from the dense feature maps in both views for  \ours.
For ``nearest" strategy, we always choose the temporally closest frame pairs from two views. 
If two randomly sampled clips temporally overlap, then we draw samples from their overlapping frames.
Otherwise, frames at the closest two ends of the two video clips are selected.
Table~\ref{tab:sampling} shows that the ``random" sampling strategy is consistently worse than the other approaches.
This can be attributed to that the temporally random sampling introduces significant noise and negatively affect the model performance.  
%
We do not see significant performance differences by using the ``center" or ``nearest" sampling strategy. 
For the simplicity, we choose the ``center" sampling strategy throughout our experiments.

\noindent \textbf{Loss endpoints.}
We analyze how global and local contrastive losses can be integrated together for vision tasks. 
In this study, we attach the proposed local loss to different endpoints of the network and analyze how it interacts with the global loss.
As shown in Fig.~\ref{fig:global_local_loss}(a) the region features are from the C5 endpoint for this model. 
For the model in Fig.~\ref{fig:global_local_loss}(b), we first perform a 2$\times$2 average pooling on the C4 feature map to reduce its spatial resolution from 14$\times$14 to 7$\times$7 and then apply the region-level loss.
For the model in Fig~\ref{fig:global_local_loss}(c), we duplicates the $res5$ block of the network and then apply the global loss on $\text{C5}_g$ branch and the region loss on the $\text{C5}_r$ branch respectively.
During the inference stage for the model in Fig~\ref{fig:global_local_loss}(c), we use feature maps from $\text{C5}_g$ and $\text{C5}_r$ for video and instance-level tasks respectively.
In Table~\ref{tab:endpoints}, we observe that by simply adding the proposed region-based contrastive loss on C4$_p$ or C5, \ours outperforms the baseline method CVRL on downstream tasks already.
By branching the $res5$ block as shown in Fig~\ref{fig:global_local_loss}(c), we achieve the best trade-off of two losses and the holistic and local representations obtain better performance gains on downstream tasks.

\noindent \textbf{Context length.}
We study the effect of contextualization by varying the number of feature maps sampled along the time axis from the target view to input to the \ours transformer head.
Different number of context length indicates the number of feature maps sampled along the time axis.
Note that when the context length is zero, the method simply degrades to the vanilla region-based contrastive learning described in Section~\ref{sec:icl}. 
Table~\ref{tab:context_length_and_overhead} shows 
our model learns better representations as the context length is increased. 
This can be attributed to that as more context features provide richer information about the target view, the model can learn a better decoder function $g(\cdot, \cdot)$, resulting in higher-quality spatio-temporal representations.
%
In Table~\ref{tab:context_length_and_overhead}, we also present the number of model parameters and FLOPs for one pass of self-supervised training.
Comparing the vanilla region-based model (second row) to the baseline model CVRL (first row), we note that the number of model parameters is increased by $74.2\%$ and the number of FLOPs is increased by $20.9\%$, where the overheads largely result from the duplicated $res5$ block.
When switching to the proposed transformer decoder with context length equals to $1$, the number of model parameters is decreased to $71.76$M while the FLOPs is increased by $0.16$B.
This is due to the fact that we use multi-head self-attention with fewer hidden units, which is more parameter efficient.
Finally, when increasing context lengths, we notice only slight increases in the number of FLOPs.

\noindent \textbf{Boxes type.}
Table~\ref{tab:box_type} shows how different location priors affect the representations learning.
We study three types of boxes generated using different methods: randomly generated boxes, boxes derived from low-level image cues and detector-generated boxes. 
Overall, our method performs equally well regardless whether region locations are accurate or not.
%
We reason that each region could be understood as an instance crop in the scene and \ours does not require the crop to be object-centric.
This observation is aligned with previous self-supervised learning methods on images~\cite{he2019moco, chen2020simclr} and videos~\cite{qian2021cvrl}.
The experiment suggests the robustness of the proposed method.

%% file: 5.Conclusion.tex
\section{Conclusion}
\label{sec:conclusion}


In this paper, we propose a novel self-supervised learning framework that facilitates learning versatile spatio-temporally fine-grained representations in videos. 
A simple architecture design is proposed to reconcile holistic and local representations learning in one single framework.
Extensive experiments are carried out to demonstrate the efficacy of the proposed method.  
%
%
In the future, we plan to experiment on more video tasks, such as video segmentation and temporal localization.
%

\noindent \textbf{Acknowledgment.} We thank Olivier Henaff (DeepMind) for providng image segmentation code; Ang Li (DeepMind) for the help with AVA-Kinetics; Xiao Zhang (U. Chicago) and Jianing Wei (Google) for the help with object tracking.

%% file: Sup_Materials/r3d_table.tex
\newlength\savewidth\newcommand\shline{\noalign{\global\savewidth\arrayrulewidth\global\arrayrulewidth 1pt}\hline\noalign{\global\arrayrulewidth\savewidth}}
\newcommand{\tablestyle}[2]{\setlength{\tabcolsep}{#1}\renewcommand{\arraystretch}{#2}\centering\footnotesize}
\newcommand{\blocks}[3]{\multirow{3}{*}{\(\left[\begin{array}{c}\text{1$\times$1$^\text{2}$, #2}\\[-.1em] \text{1$\times$3$^\text{2}$, #2}\\[-.1em] \text{1$\times$1$^\text{2}$, #1}\end{array}\right]\)$\times$#3}}
\newcommand{\blockt}[3]{\multirow{3}{*}{\(\left[\begin{array}{c}\text{\underline{3$\times$1$^\text{2}$}, #2}\\[-.1em] \text{1$\times$3$^\text{2}$, #2}\\[-.1em] \text{1$\times$1$^\text{2}$, #1}\end{array}\right]\)$\times$#3}}

\begin{table}[h]
    \centering
    \resizebox{\columnwidth}{!}{
        \tablestyle{5pt}{1.1}
        \begin{tabular}{c|c|c|c}
            \shline
            Stage & Network & Input from & Output size $T \times S^2$ \\
            \shline
            raw clip & - & - & $32\times224^\text{2}$ \\
            \hline
            data & stride \textbf{2}, 1$^\text{2}$ & raw clip & $16\times224^\text{2}$ \\
            \hline
            \multirow{2}{*}{res1} &
            \multicolumn{1}{c|}{\underline{$\textbf{5}\times7^\text{2}$}, {64}} &
            \multirow{2}{*}{data} &
            \multirow{2}{*}{$8\times112^\text{2}$}  \\
            & stride \textbf{2}, 2$^\text{2}$ & \\
            \hline
            \multirow{2}{*}{pool1} &
            \multicolumn{1}{c|}{$1\times3^\text{2}$ max} & 
            \multirow{2}{*}{res1} &
            \multirow{2}{*}{$8\times56^\text{2}$} \\
            & stride 1, 2$^\text{2}$ & \\
            \hline
            \multirow{3}{*}{res2} & \blocks{{256}}{{64}}{3} &
            \multirow{3}{*}{pool1} &
            \multirow{3}{*}{$8\times56^\text{2}$}  \\
            &  & \\
            &  & \\
            \hline
            \multirow{3}{*}{res3} & \blocks{{512}}{{128}}{4} &
            \multirow{3}{*}{res2} & 
            \multirow{3}{*}{$8\times28^\text{2}$}  \\
            &  & \\
            &  & \\
            \hline
            \multirow{3}{*}{res4} & \blockt{{1024}}{{256}}{6} & 
            \multirow{2}{*}{res3} &
            \multirow{3}{*}{$8\times14^\text{2}$} \\
            &  & \\
            &  & \\
            \hline
            \multirow{3}{*}{res5$_{r}$} & \blockt{{2048}}{{512}}{3} &
            \multirow{2}{*}{res4} &
            \multirow{3}{*}{$8\times7^\text{2}$} \\
            &  & \\
            &  & \\
            \hline
            \multirow{3}{*}{res5$_g$} & \blockt{{2048}}{{512}}{3} &
            \multirow{2}{*}{res4} &
            \multirow{3}{*}{$8\times7^\text{2}$} \\
            &  & \\
            &  & \\
			\shline
    \end{tabular}
    }
    \vspace{2mm}
    \caption{\textbf{Base network $\boldsymbol{f(\cdot)}$: a ResNet3D-50 (R3D-50) based encoder.} 
    }
    \label{tab:r3d_network}
\vspace{-4mm}
\end{table}

%% file: Sup_Materials/tx_table.tex

\begin{table}[t]
    \centering
    \resizebox{\columnwidth}{!}{
        \tablestyle{5pt}{1.1}
        \begin{tabular}{c|c|c|c}
            \shline
            Stage & Input, Dimension & Network & Output \\
            \shline
            Linear project & $\boldsymbol{h}$, N$\times$2048 & n\_nodes=128 & Query \\
            \hline
            Linear project & $\boldsymbol{F_c^\prime}$, M$\times$2048 & n\_nodes=128 &  Key \\
            \hline
            Linear project & $\boldsymbol{F_c^\prime}$, M$\times$2048 & n\_nodes=128 &  Value \\
            \hline
            \multirow{3}{*}{MHSA} & Query, N$\times$128 & hidden\_size=128 & \multirow{3}{*}{Hidden} \\
            & Key, M$\times$128 & n\_heads=3 & \\
            & Value, M$\times$128 & n\_layers=3 & \\
            \hline
            Linear project & Hidden, N$\times$128 & n\_nodes=2048 & $\boldsymbol{z}$ \\
			\shline
    \end{tabular}
    }
    \vspace{2mm}
    \caption{\textbf{\ours head $\boldsymbol{g(\cdot, \cdot)}$: a transformer-based decoder.} 
    The inputs are the region features $\boldsymbol{h}$ and the context features $\boldsymbol{F_c^\prime}$ and the outputs are the transformed features $\boldsymbol{z}$. N and M are the number of tokens of $\boldsymbol{h}$ and $\boldsymbol{F_c^\prime}$ respectively.}
    \label{tab:tx_network}
\vspace{-4mm}
\end{table}

%% file: Sup_Materials/attention.tex
\begin{figure*}[t!]
\centering
\setlength\tabcolsep{1pt} 
\begin{tabular}{cccc}
  Source &
  Thumbnail & 
  Target &
  Attention
\\
 {\includegraphics[width=0.23\linewidth]{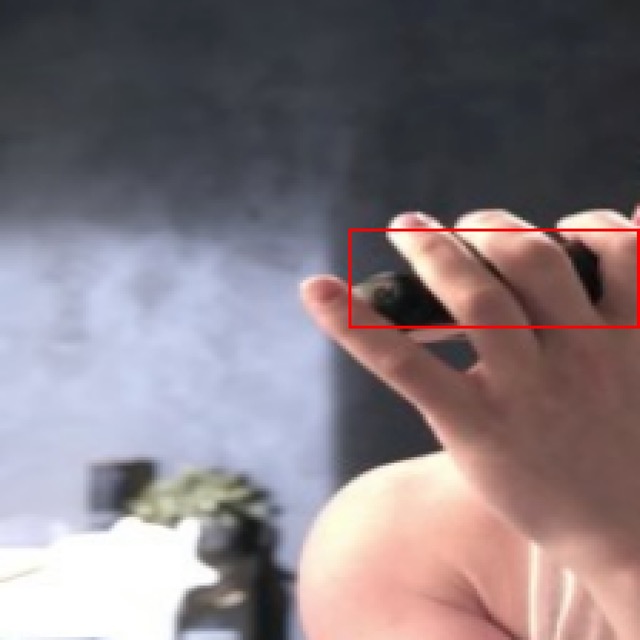}} &
 {\includegraphics[width=0.23\linewidth]{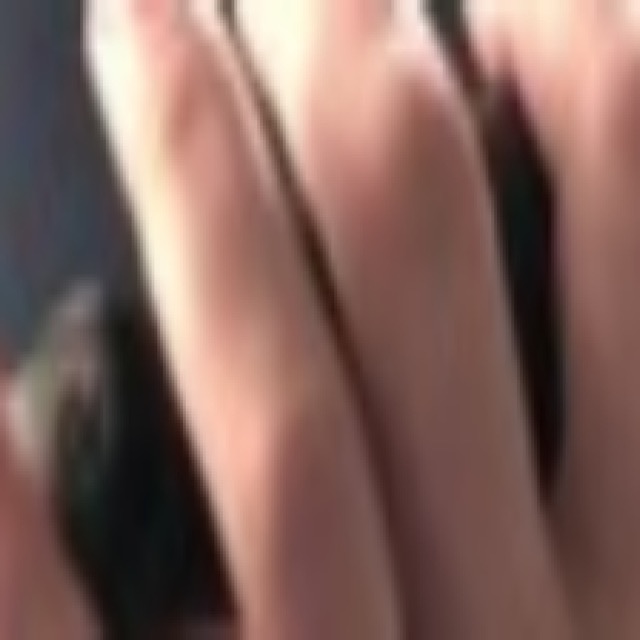}} &
 {\includegraphics[width=0.23\linewidth]{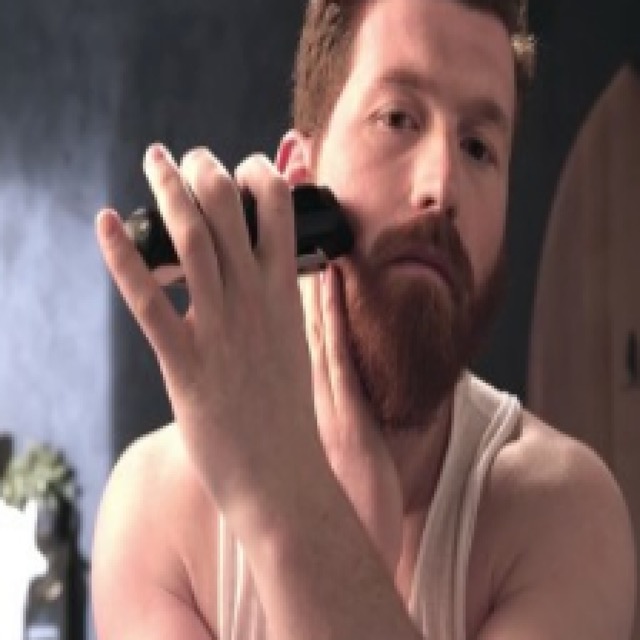}} &
 {\includegraphics[width=0.23\linewidth]{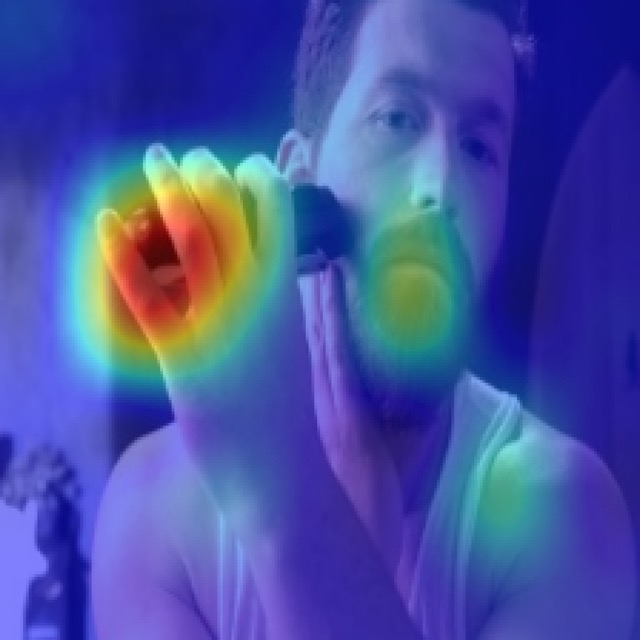}} 
 \\ 
 {\includegraphics[width=0.23\linewidth]{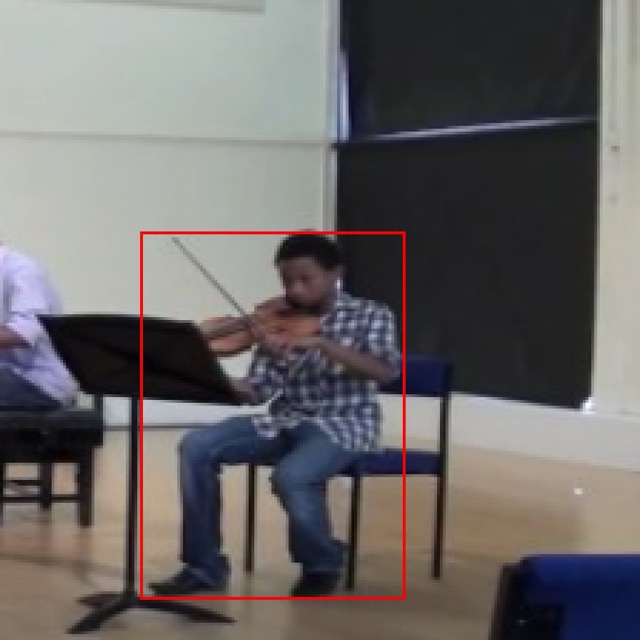}} &
 {\includegraphics[width=0.23\linewidth]{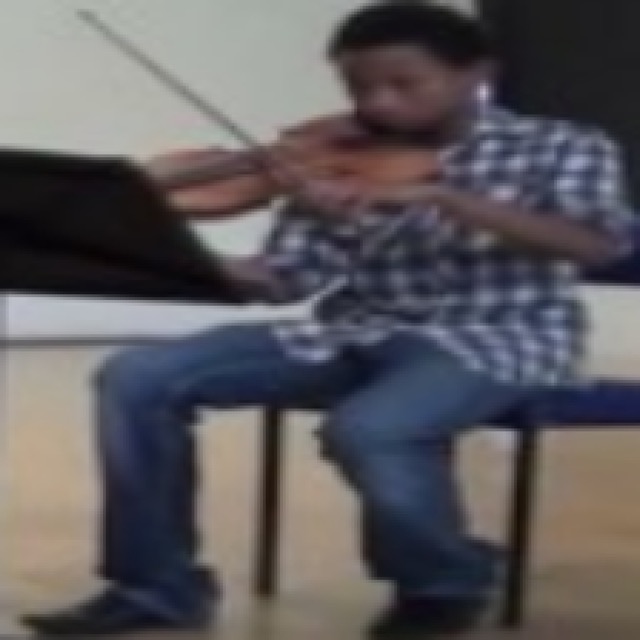}} &
 {\includegraphics[width=0.23\linewidth]{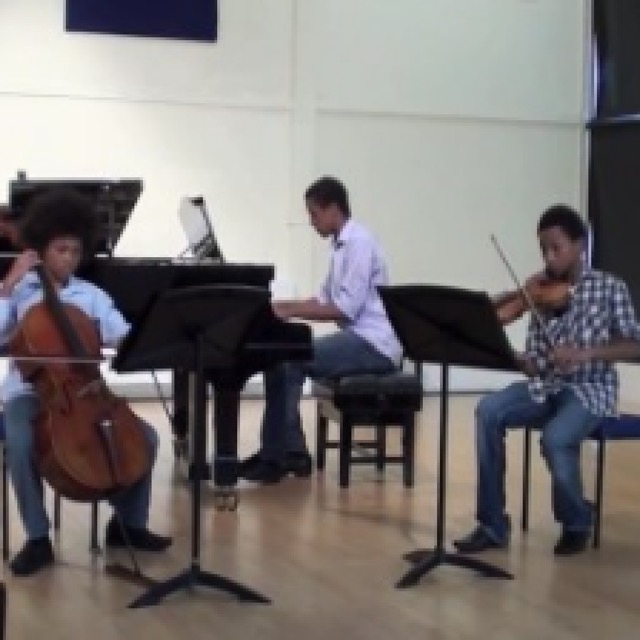}} &
 {\includegraphics[width=0.23\linewidth]{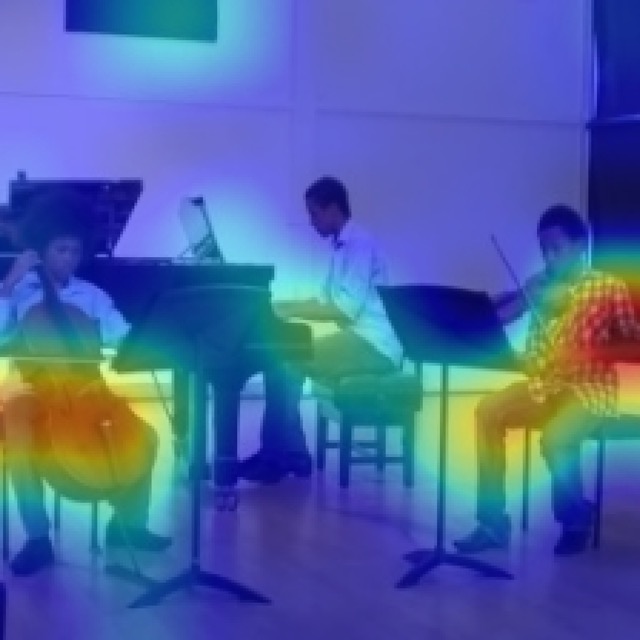}} 
 \\ 
 {\includegraphics[width=0.23\linewidth]{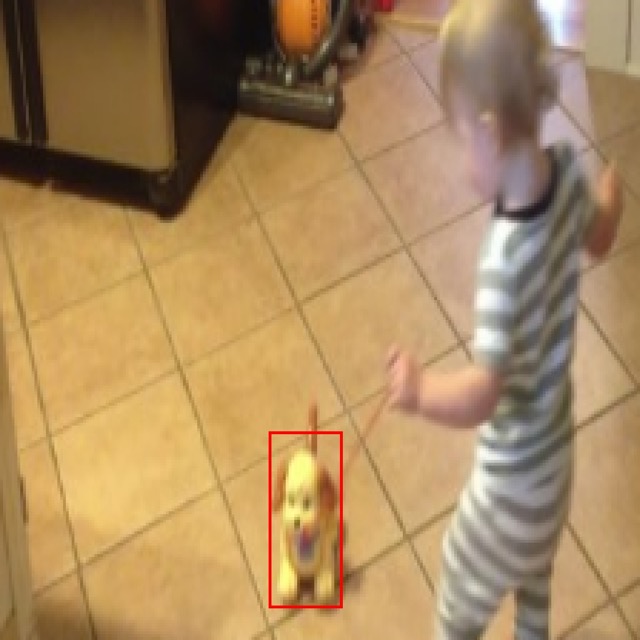}} &
 {\includegraphics[width=0.23\linewidth]{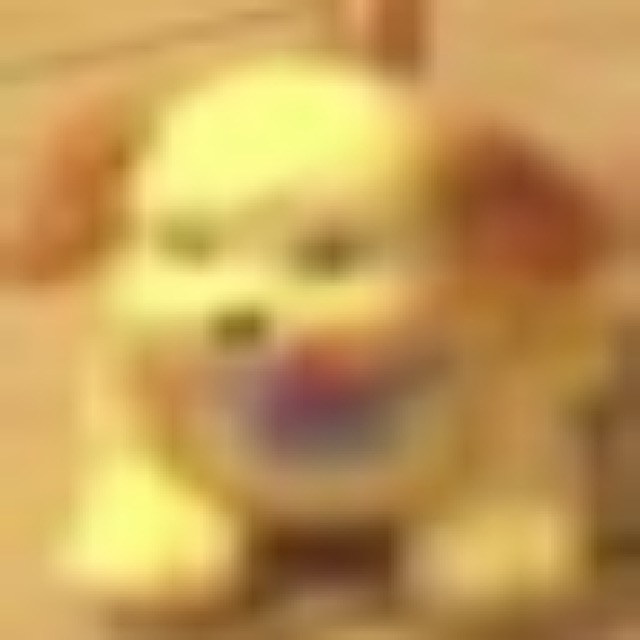}} &
 {\includegraphics[width=0.23\linewidth]{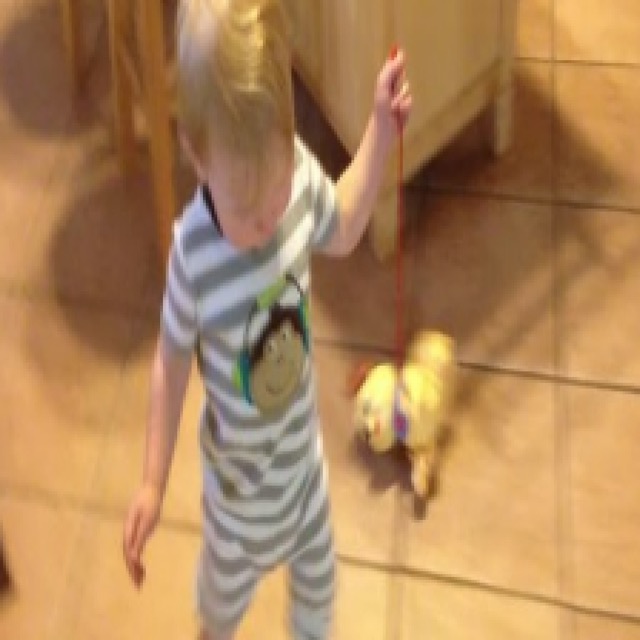}} &
 {\includegraphics[width=0.23\linewidth]{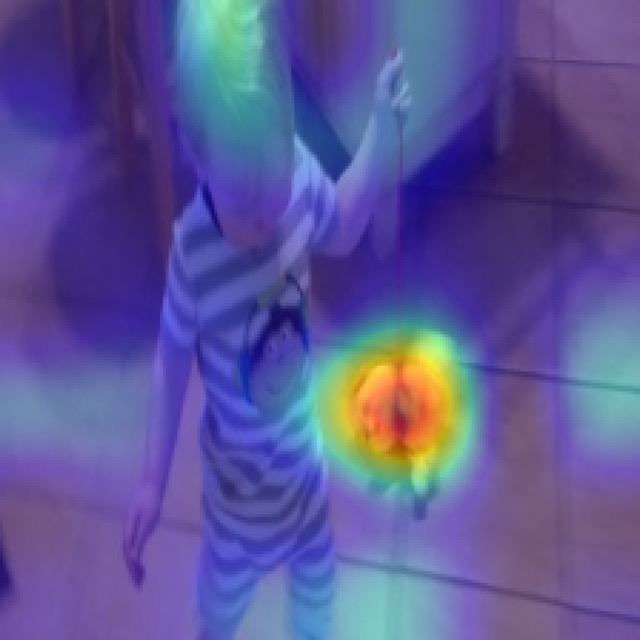}} 
 \\ 
 {\includegraphics[width=0.23\linewidth]{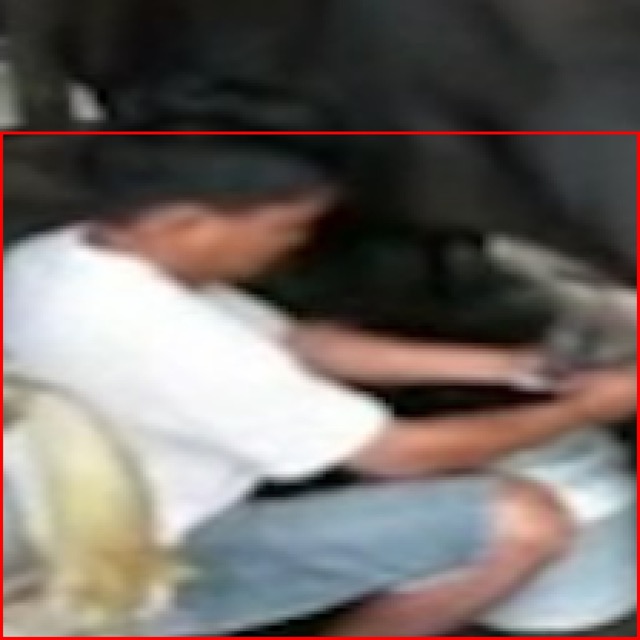}} &
 {\includegraphics[width=0.23\linewidth]{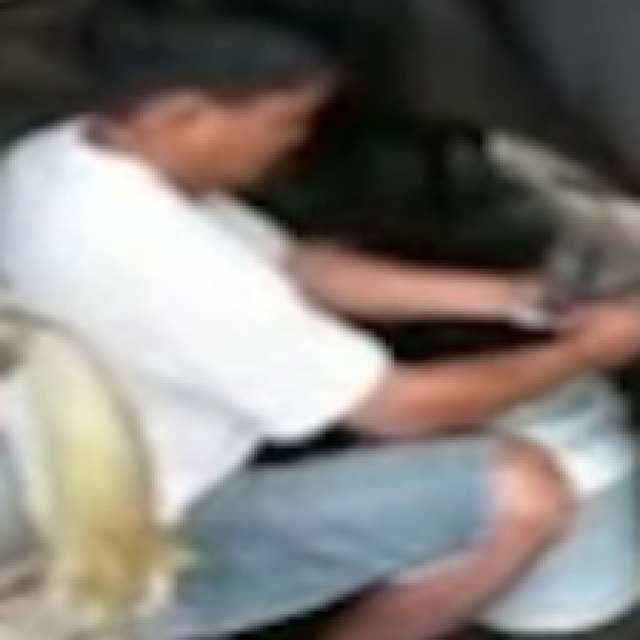}} &
 {\includegraphics[width=0.23\linewidth]{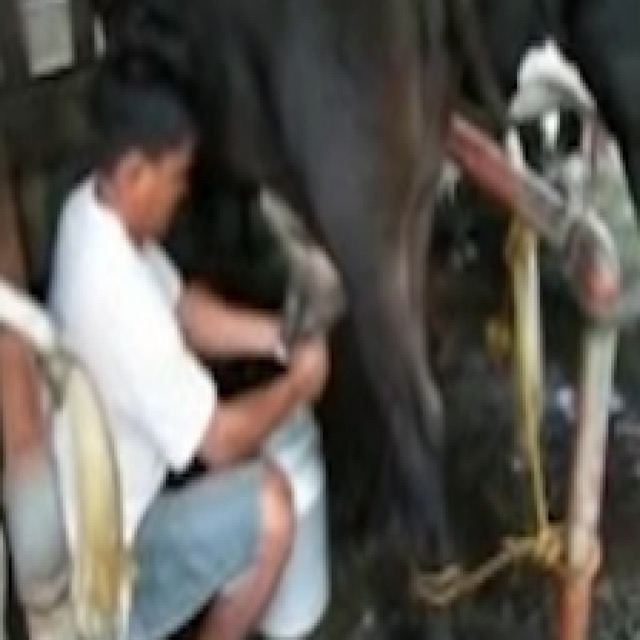}} &
 {\includegraphics[width=0.23\linewidth]{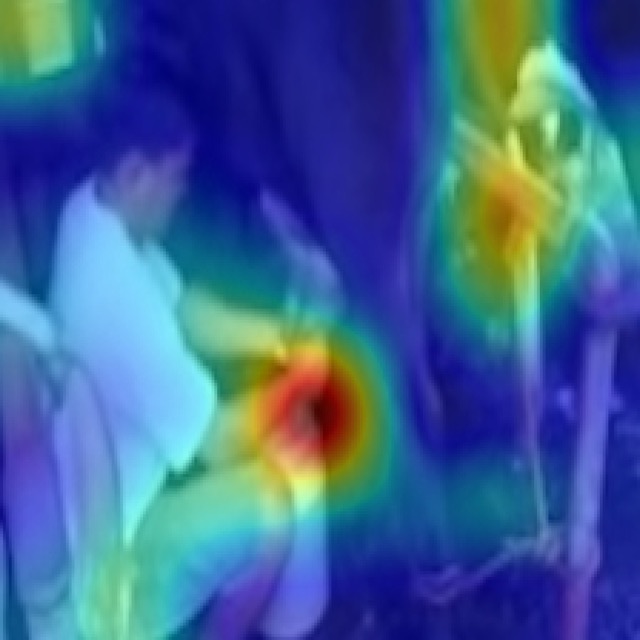}} 
 \\ 
 {\includegraphics[width=0.23\linewidth]{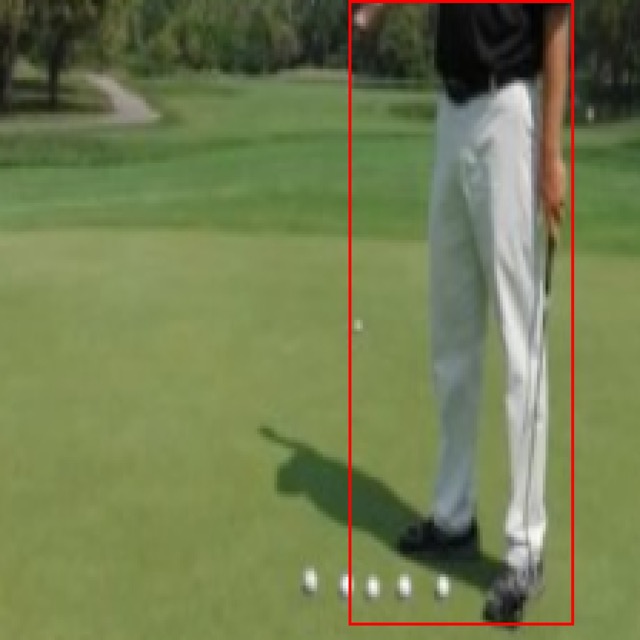}} &
 {\includegraphics[width=0.23\linewidth]{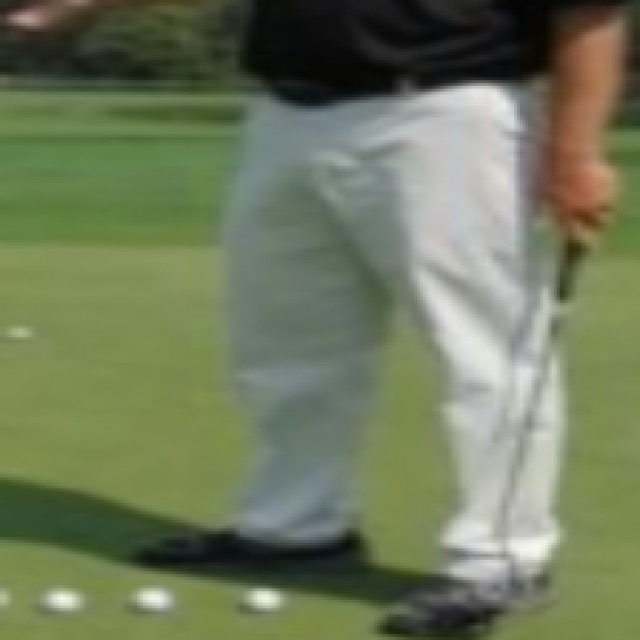}} &
 {\includegraphics[width=0.23\linewidth]{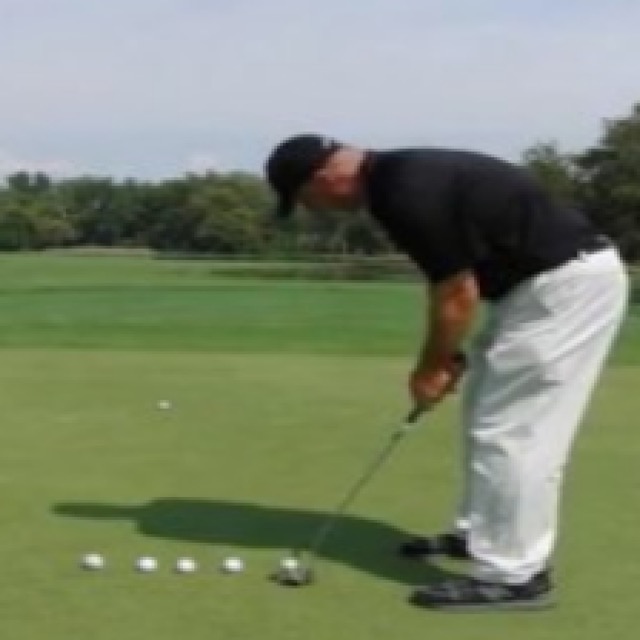}} &
 {\includegraphics[width=0.23\linewidth]{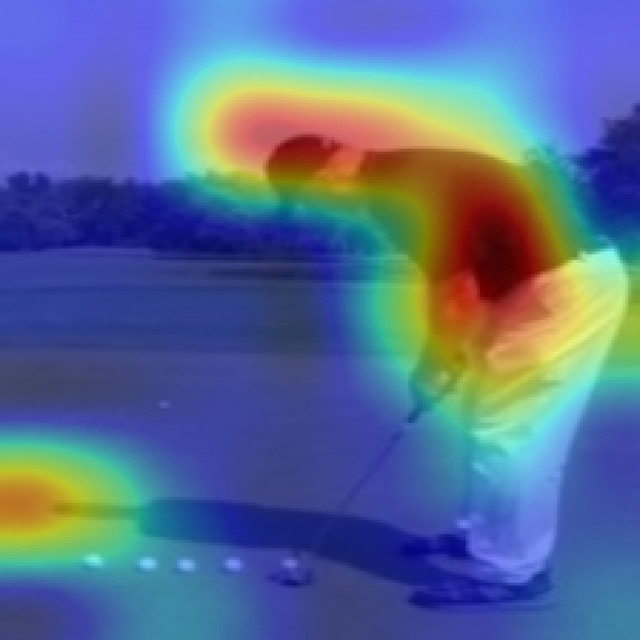}} 
 
  \end{tabular}
  \centering
  \caption{
  \textbf{Visualization of the attention during the training.}
  We use boxes from the object detector to pool region features from the source view as the query in order to generate the attention maps given the context (features from the target view).
  Interestingly, the model learns to attend to not only the corresponding instance in the target frame, but also to some other semantically meaningful objects the instance potentially interacts with.
  }
  \vspace{-15pt}
  \label{fig:attention}
\end{figure*}

%% file: Sup_Materials/tracker.tex
\begin{figure*}[t!]
\centering
\setlength\tabcolsep{1pt} 
\begin{tabular}{c|ccccc}
  Input & \multicolumn{5}{c}{Prediction}
\\
 {\includegraphics[width=0.16\linewidth]{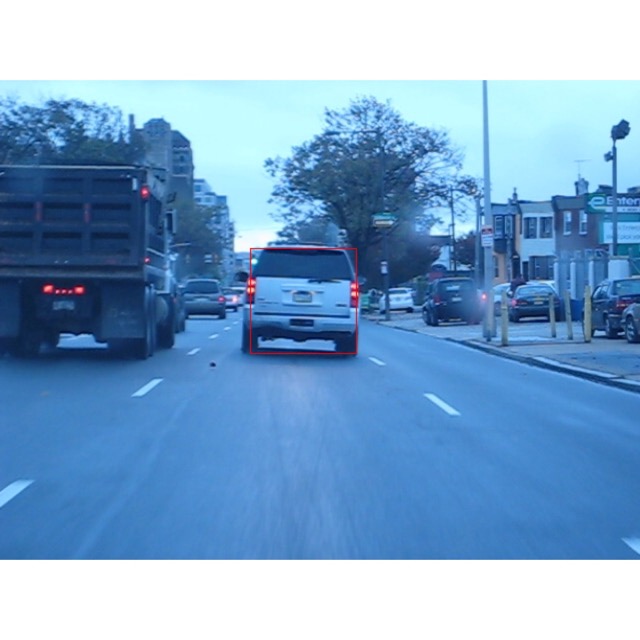}} &
 {\includegraphics[width=0.16\linewidth]{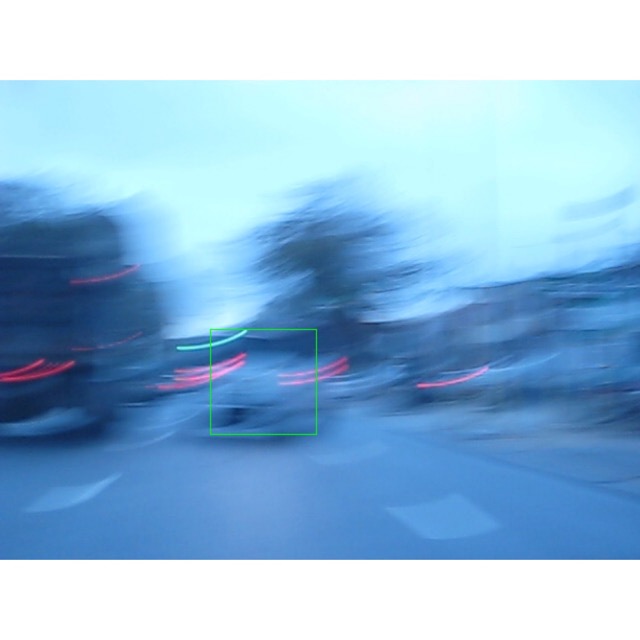}} &
 {\includegraphics[width=0.16\linewidth]{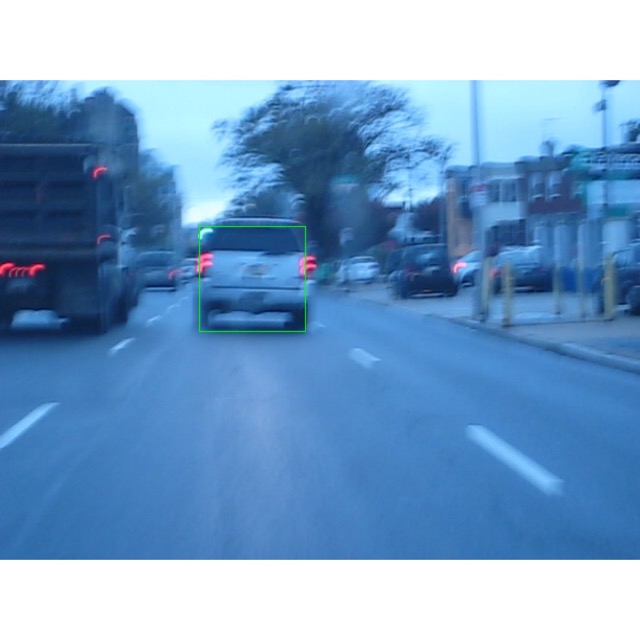}} &
 {\includegraphics[width=0.16\linewidth]{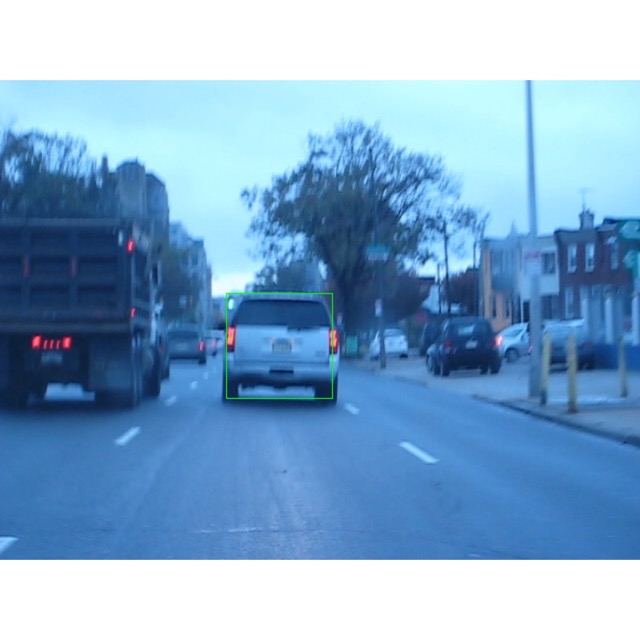}} &
 {\includegraphics[width=0.16\linewidth]{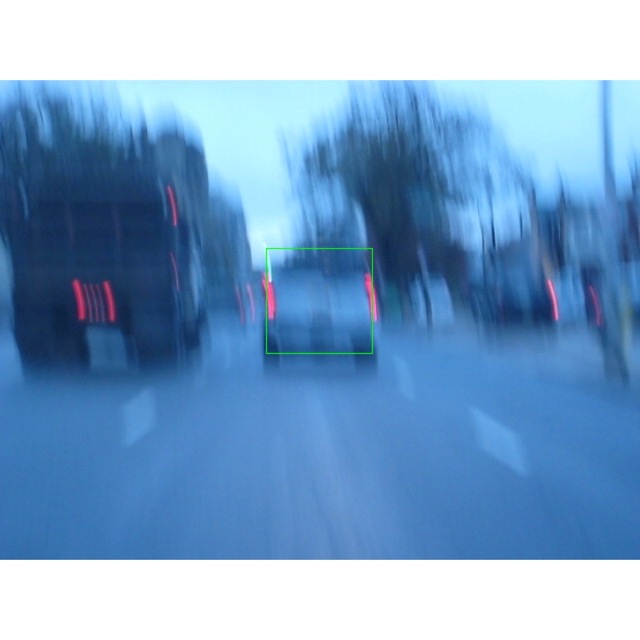}} &
 {\includegraphics[width=0.16\linewidth]{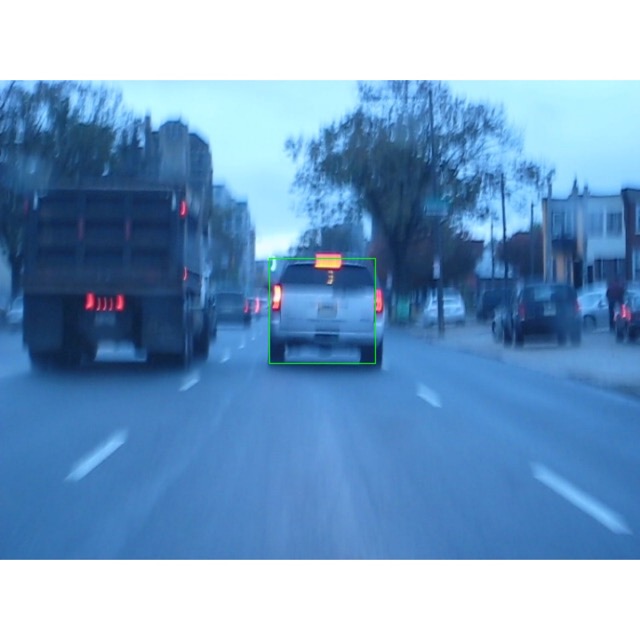}} \\
 
 {\includegraphics[width=0.16\linewidth]{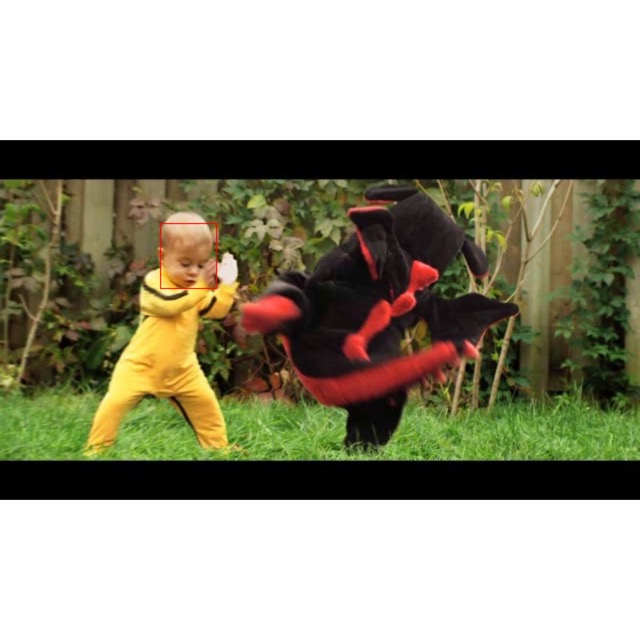}} &
 {\includegraphics[width=0.16\linewidth]{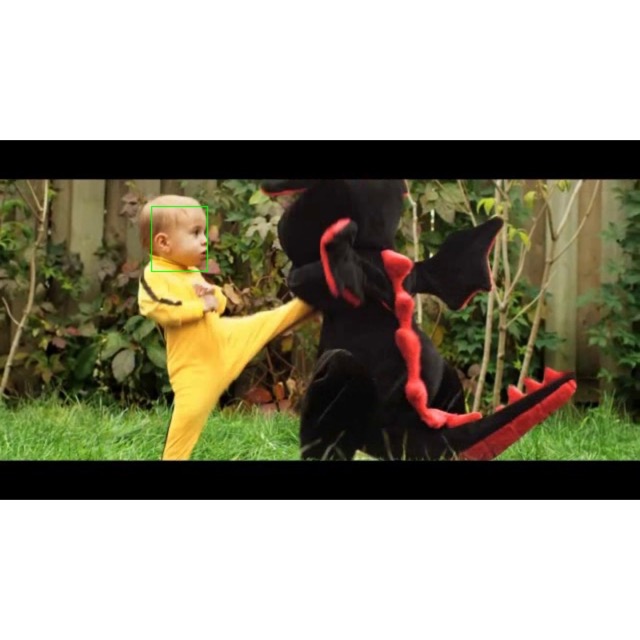}} &
 {\includegraphics[width=0.16\linewidth]{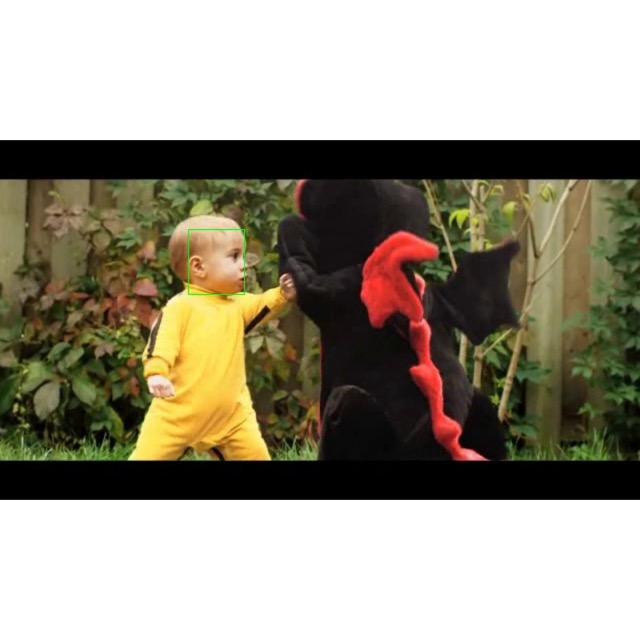}} &
 {\includegraphics[width=0.16\linewidth]{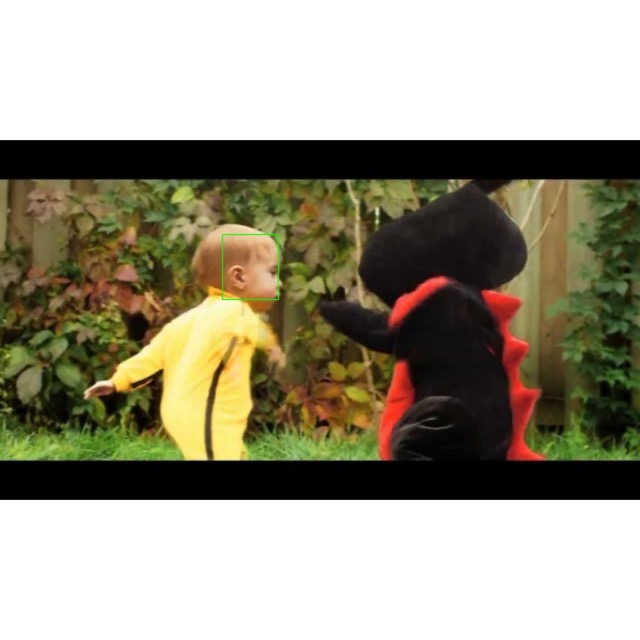}} &
 {\includegraphics[width=0.16\linewidth]{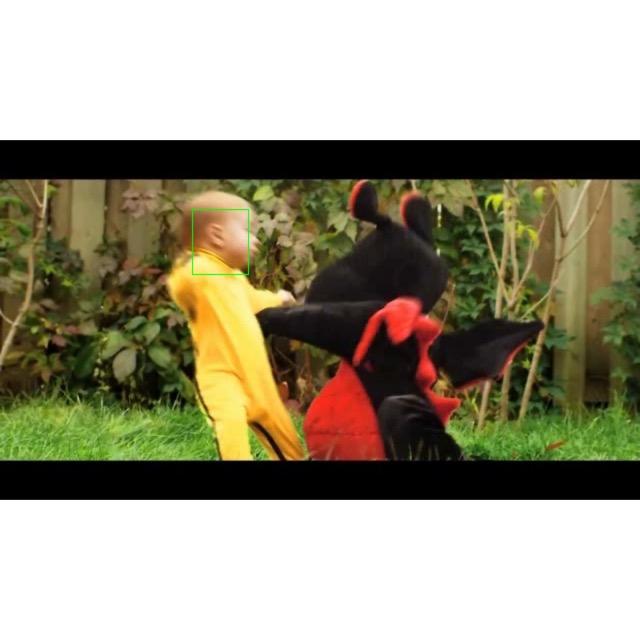}} &
 {\includegraphics[width=0.16\linewidth]{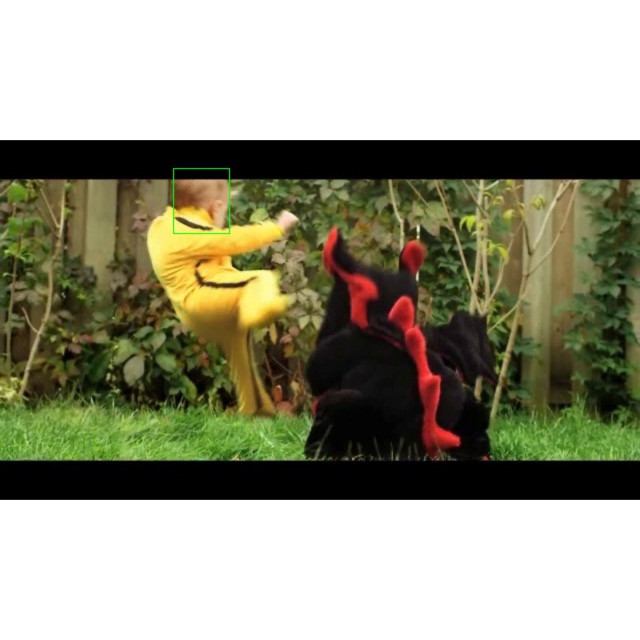}} \\
 
 {\includegraphics[width=0.16\linewidth]{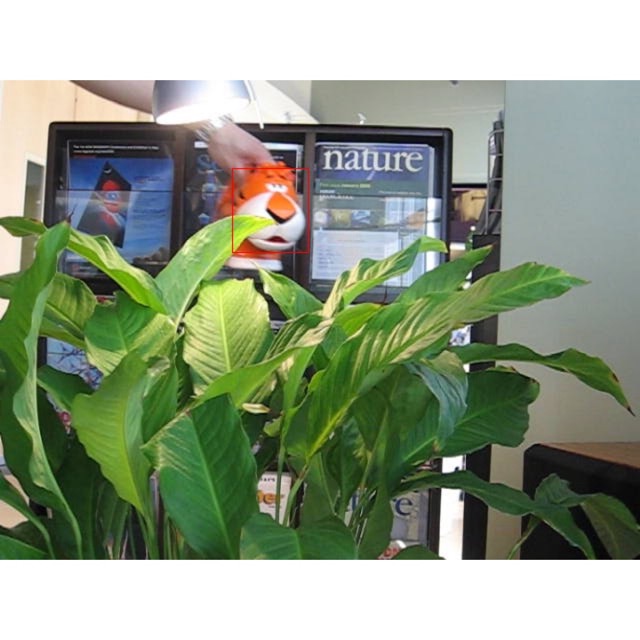}} &
 {\includegraphics[width=0.16\linewidth]{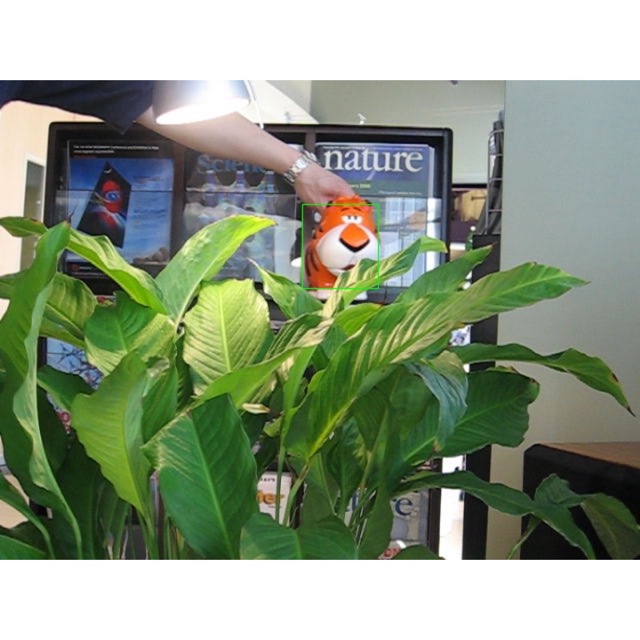}} &
 {\includegraphics[width=0.16\linewidth]{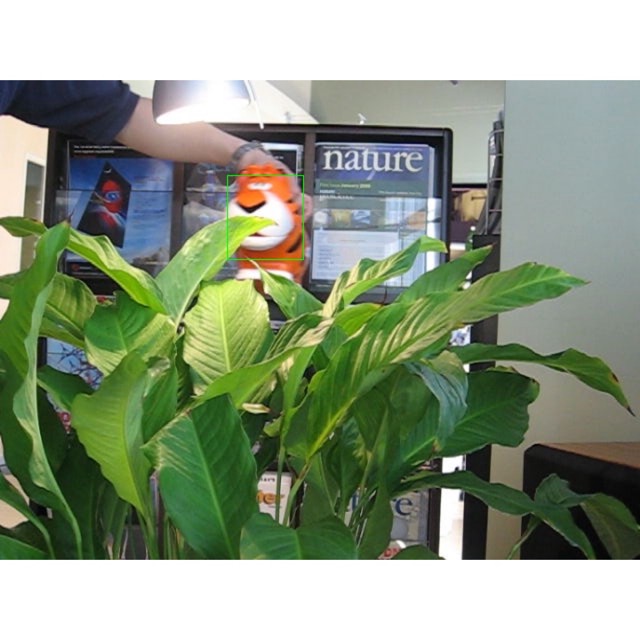}} &
 {\includegraphics[width=0.16\linewidth]{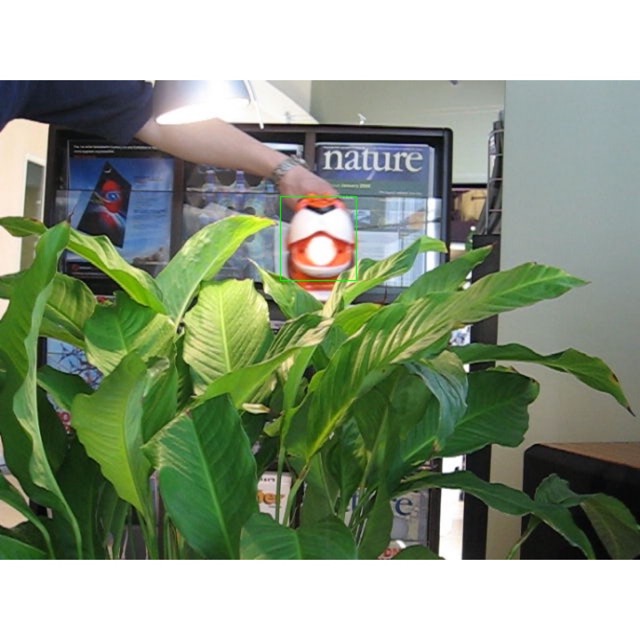}} &
 {\includegraphics[width=0.16\linewidth]{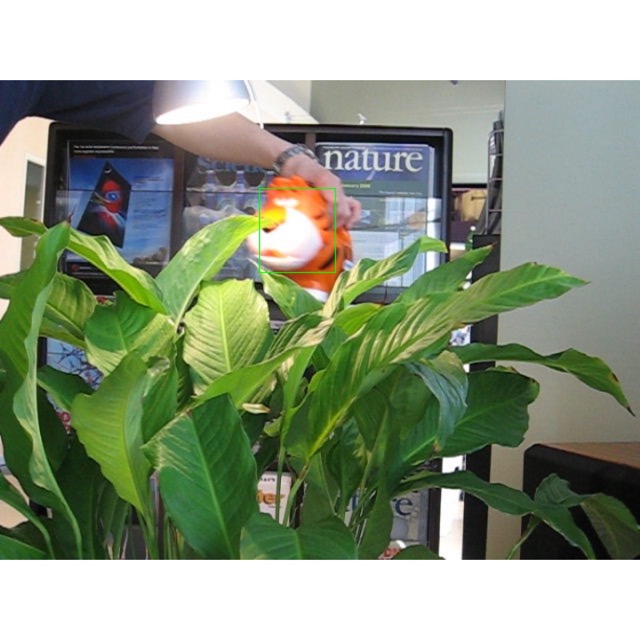}} &
 {\includegraphics[width=0.16\linewidth]{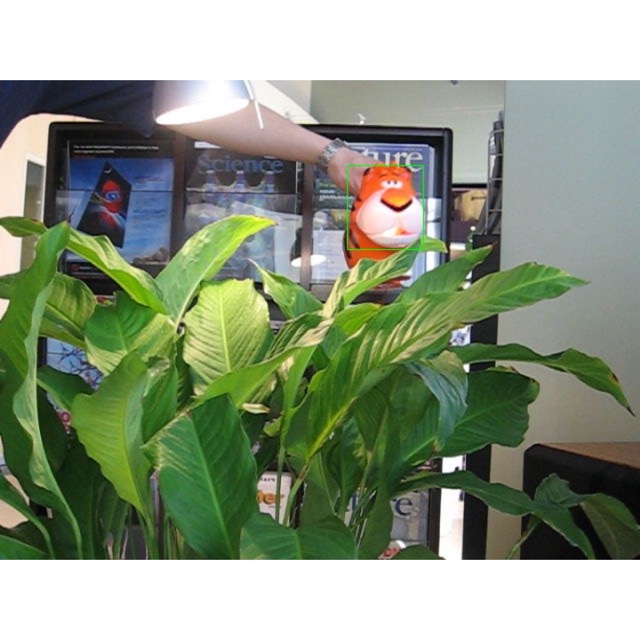}} \\
 
 {\includegraphics[width=0.16\linewidth]{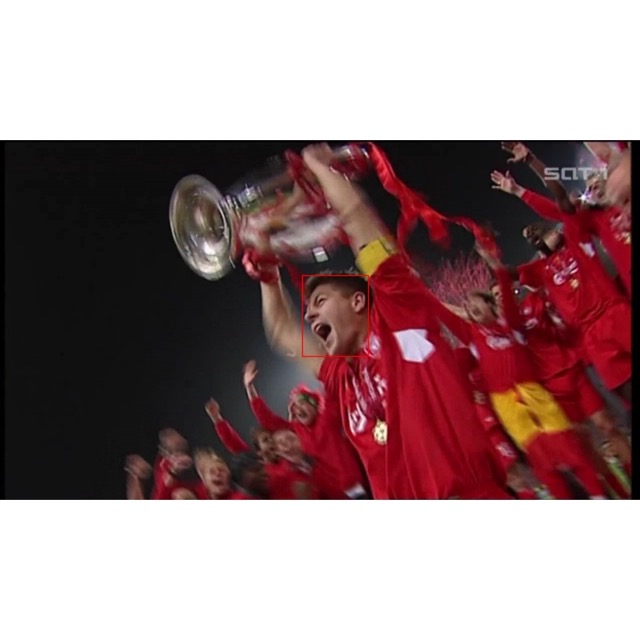}} &
 {\includegraphics[width=0.16\linewidth]{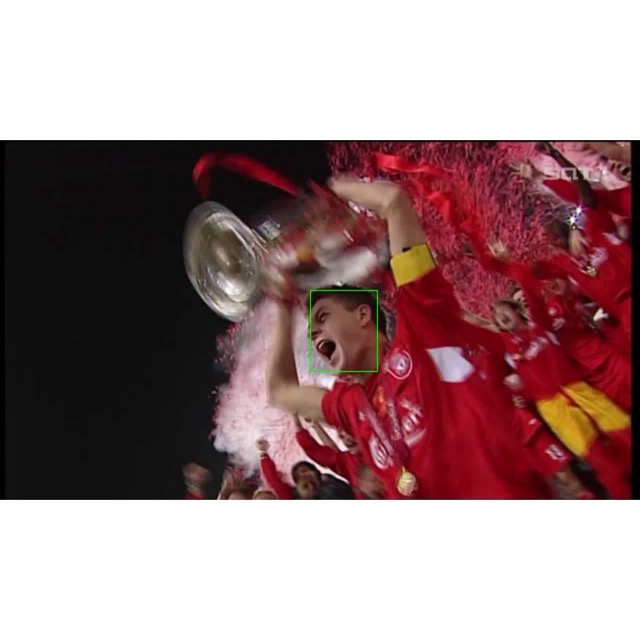}} &
 {\includegraphics[width=0.16\linewidth]{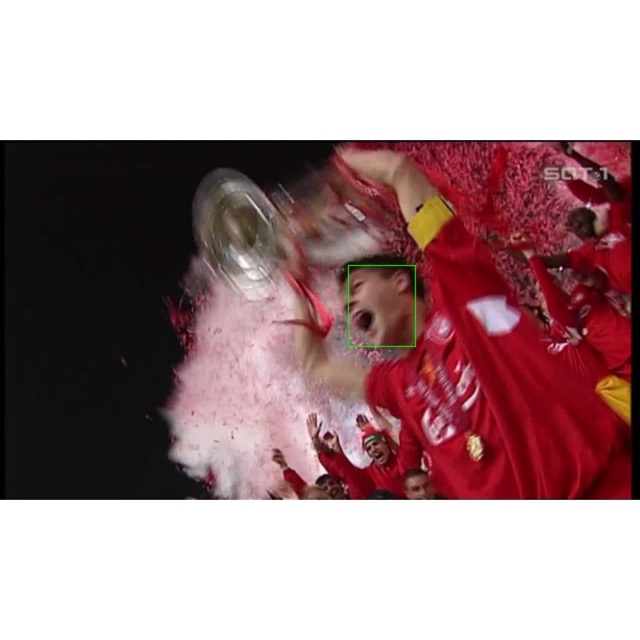}} &
 {\includegraphics[width=0.16\linewidth]{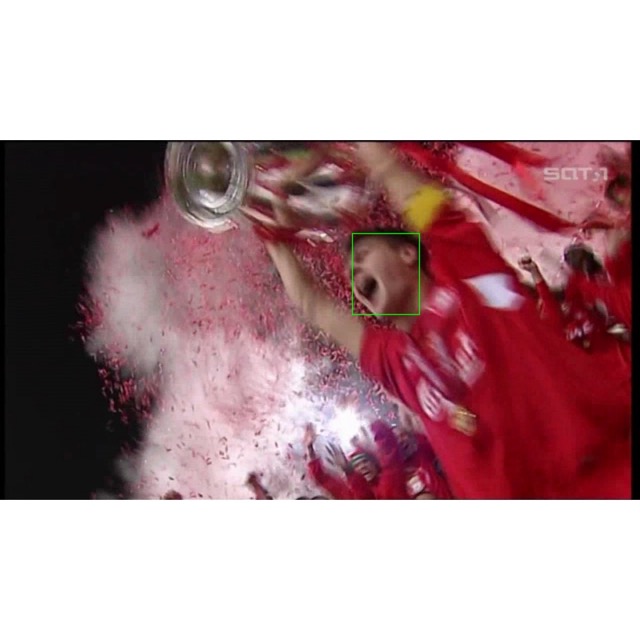}} &
 {\includegraphics[width=0.16\linewidth]{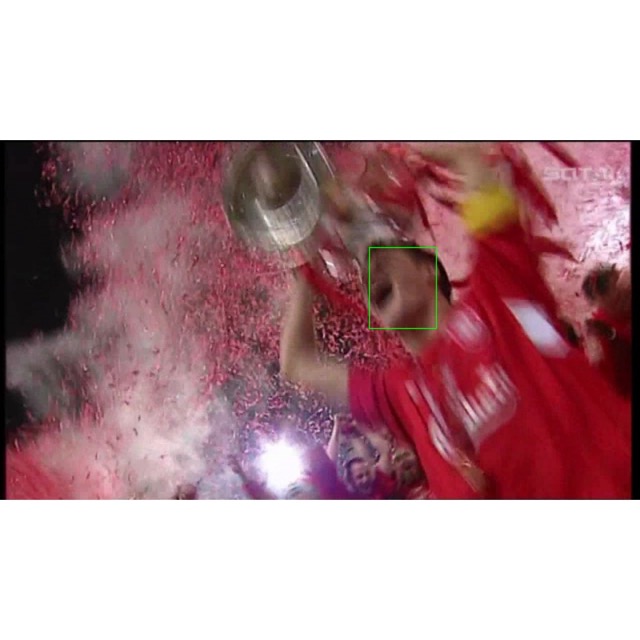}} &
 {\includegraphics[width=0.16\linewidth]{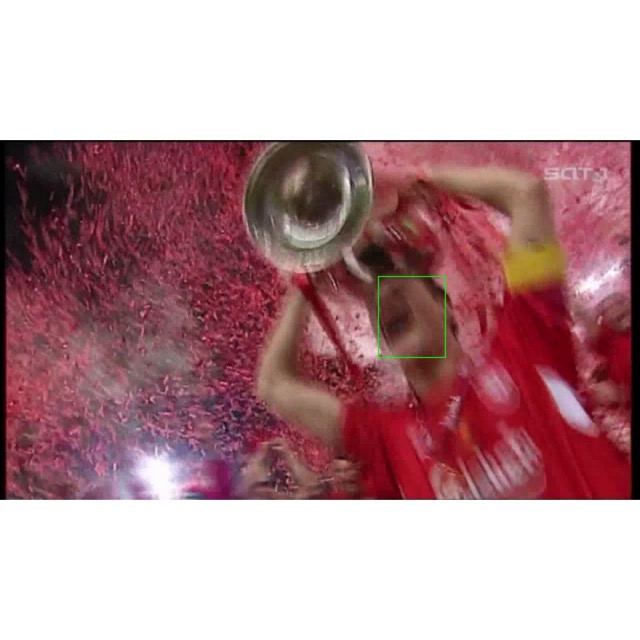}} \\
 
 {\includegraphics[width=0.16\linewidth]{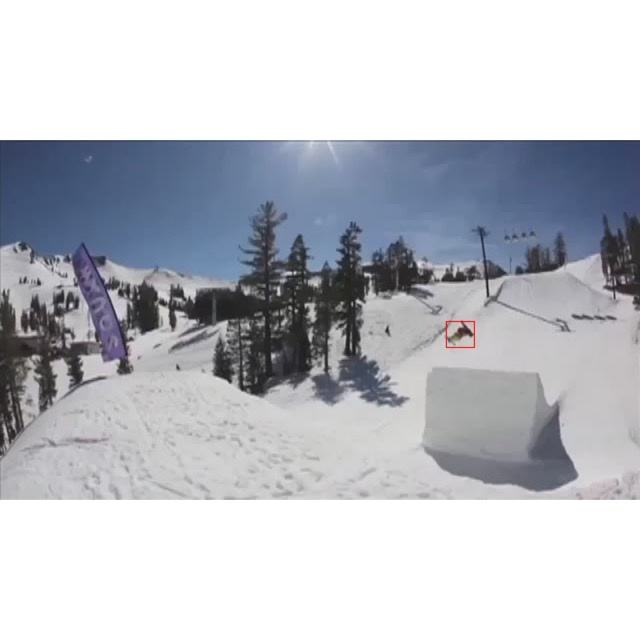}} &
 {\includegraphics[width=0.16\linewidth]{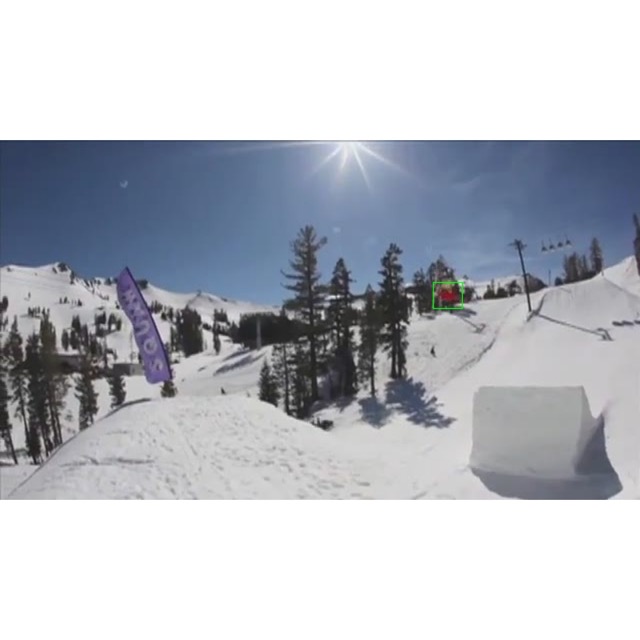}} &
 {\includegraphics[width=0.16\linewidth]{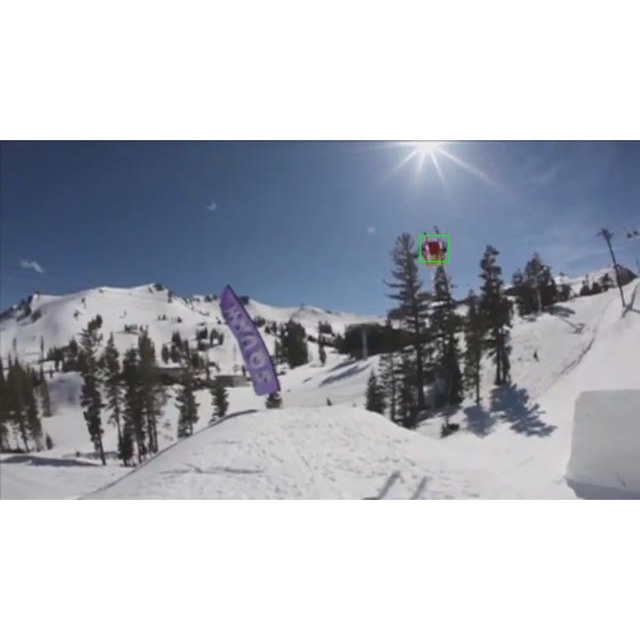}} &
 {\includegraphics[width=0.16\linewidth]{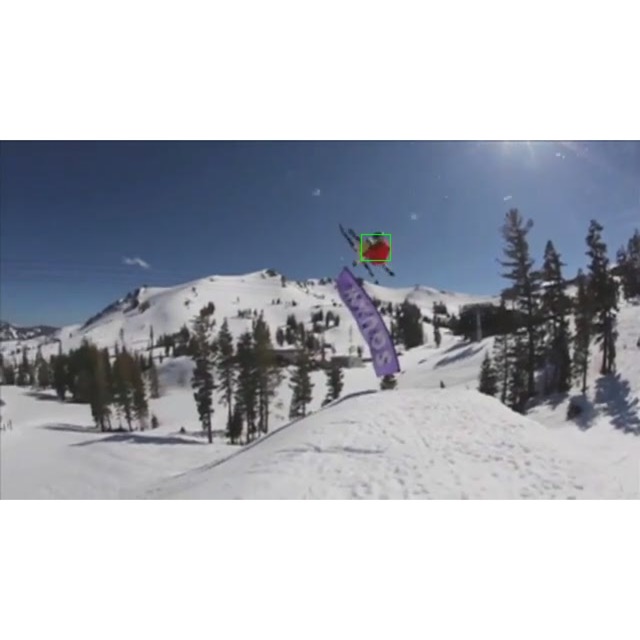}} &
 {\includegraphics[width=0.16\linewidth]{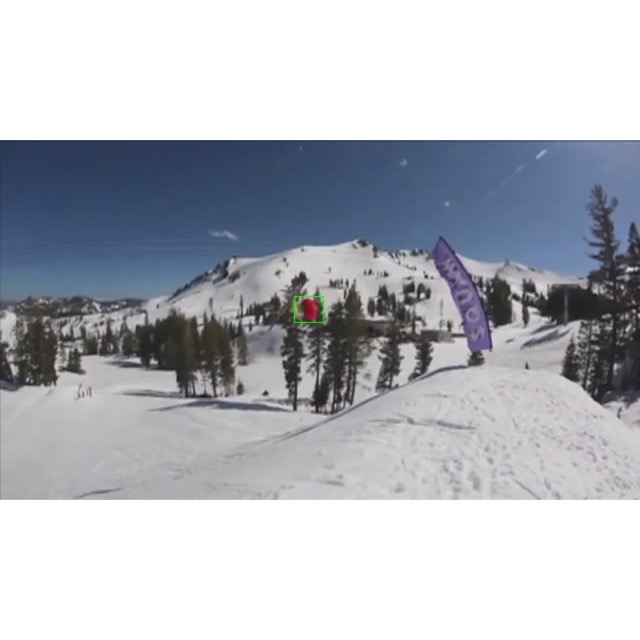}} &
 {\includegraphics[width=0.16\linewidth]{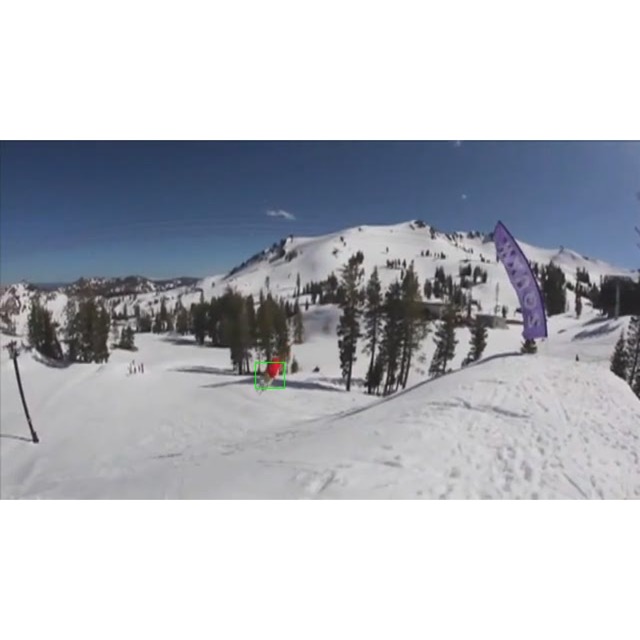}} \\
  
 {\includegraphics[width=0.16\linewidth]{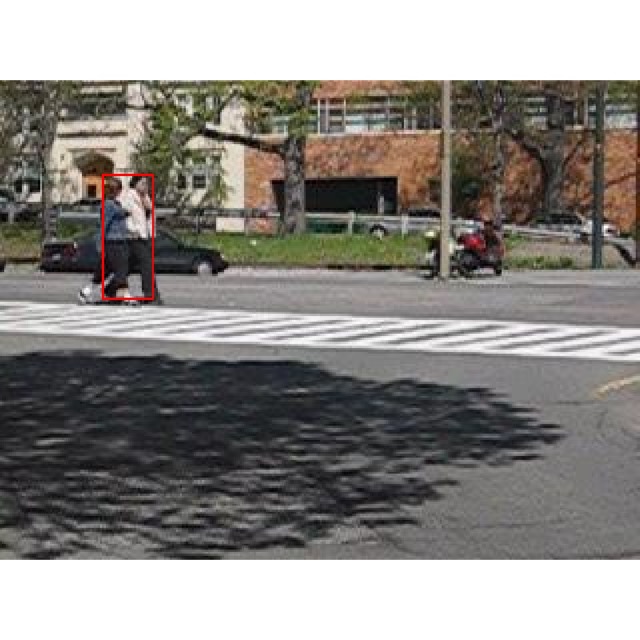}} &
 {\includegraphics[width=0.16\linewidth]{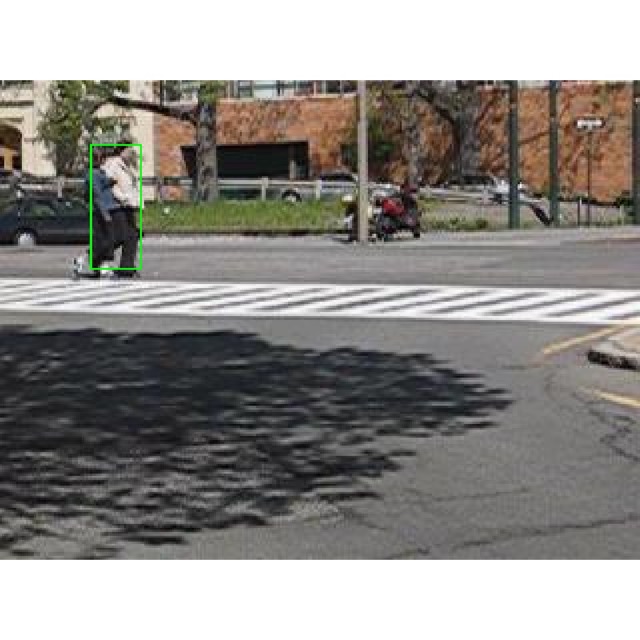}} &
 {\includegraphics[width=0.16\linewidth]{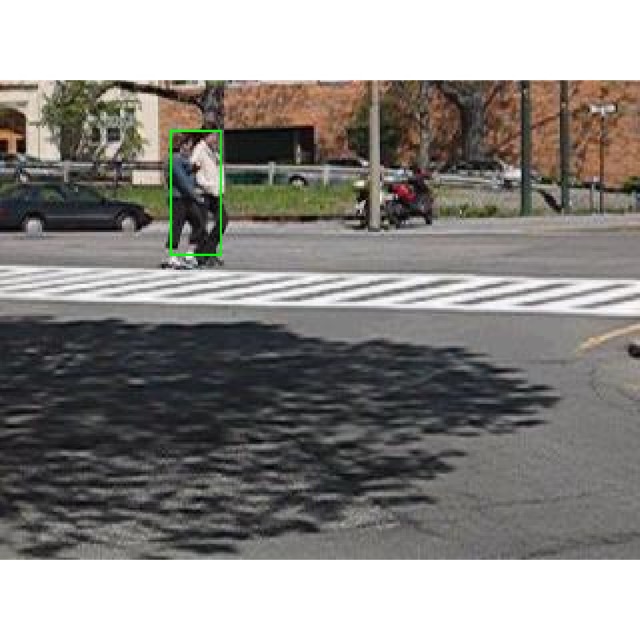}} &
 {\includegraphics[width=0.16\linewidth]{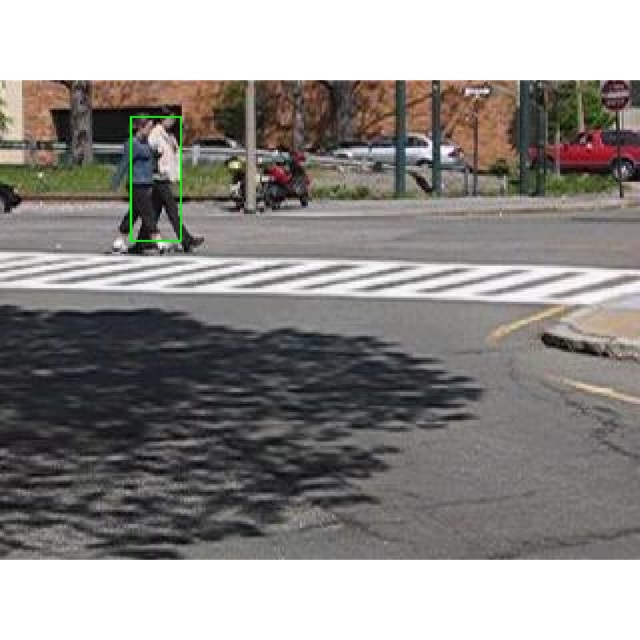}} &
 {\includegraphics[width=0.16\linewidth]{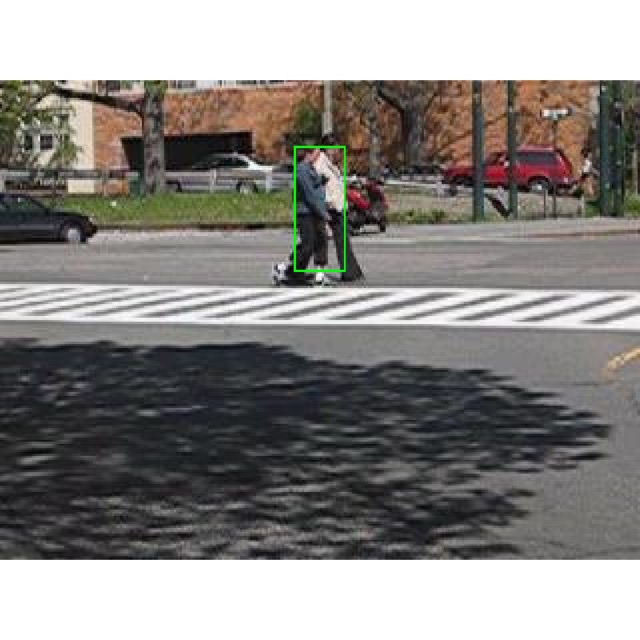}} &
 {\includegraphics[width=0.16\linewidth]{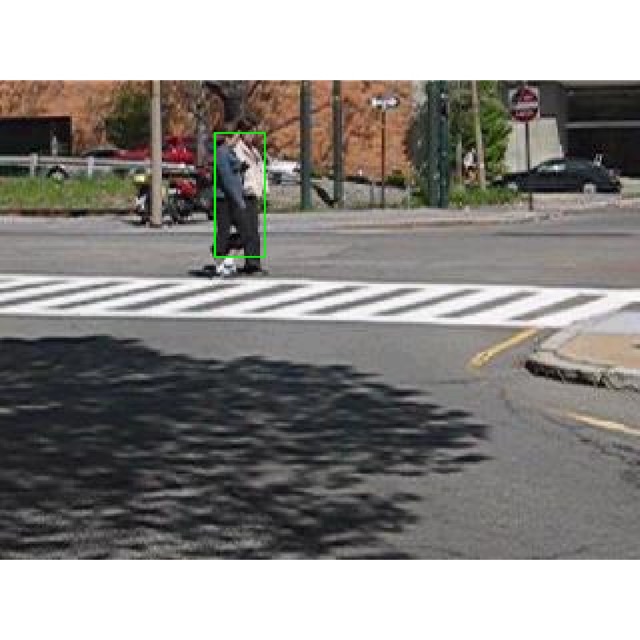}} \\
 
 {\includegraphics[width=0.16\linewidth]{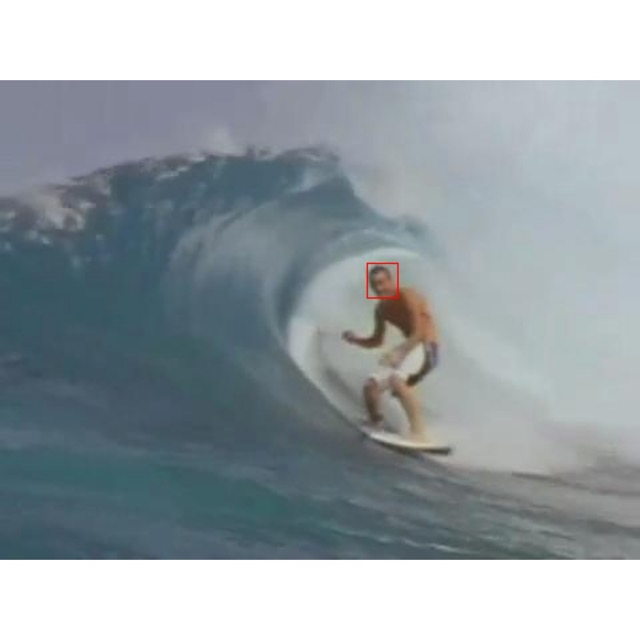}} &
 {\includegraphics[width=0.16\linewidth]{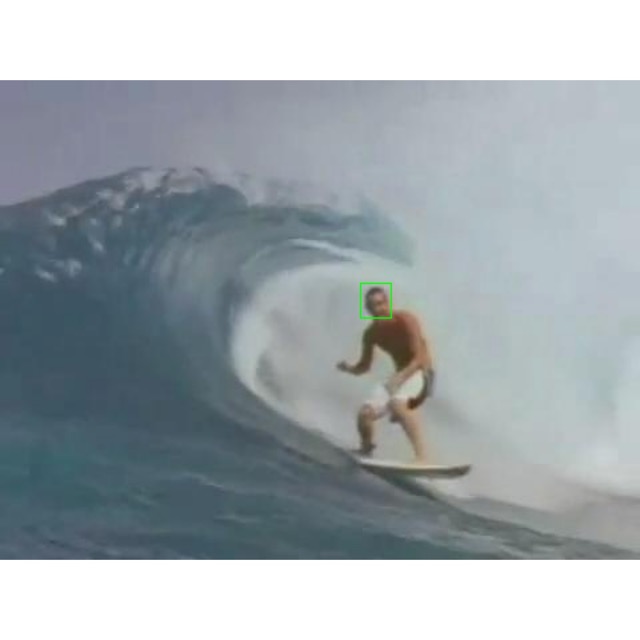}} &
 {\includegraphics[width=0.16\linewidth]{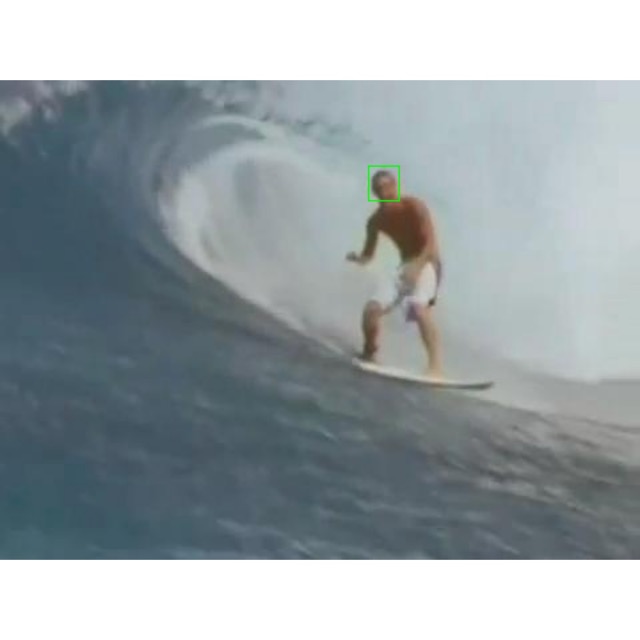}} &
 {\includegraphics[width=0.16\linewidth]{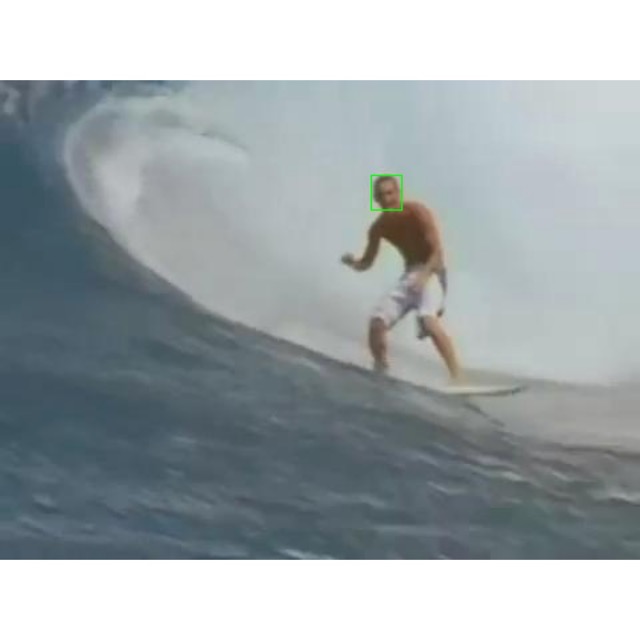}} &
 {\includegraphics[width=0.16\linewidth]{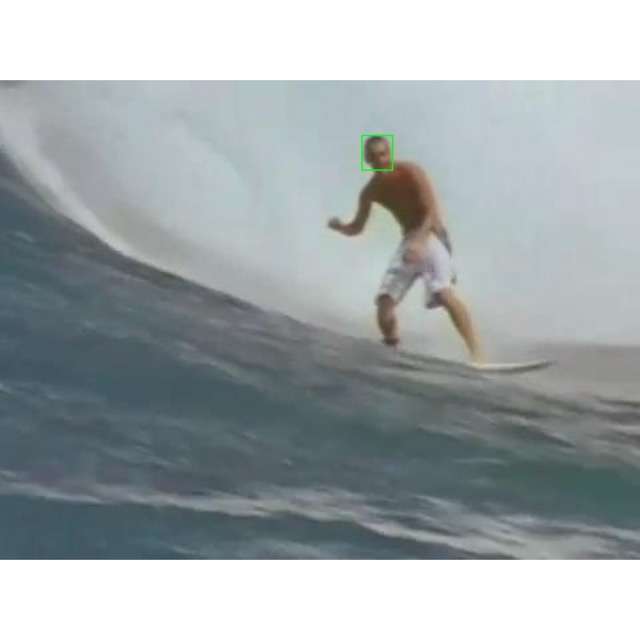}} &
 {\includegraphics[width=0.16\linewidth]{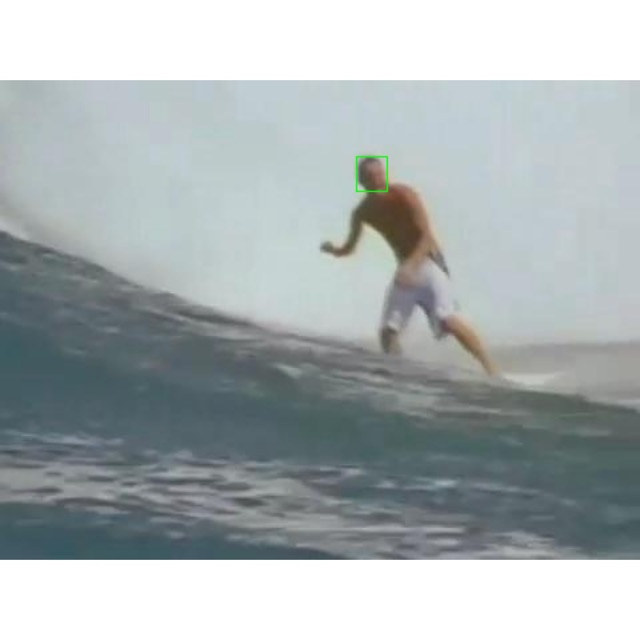}} \\
 
  \end{tabular}
  \centering
  \caption{ 
  Qualitative results for visual object tracking on OTB2015~\cite{wu2013otb}. Best view in color.
  }
  \label{fig:tracker}
\end{figure*}